\documentclass[10pt,journal,compsoc]{IEEEtran}
\ifCLASSOPTIONcompsoc
  \usepackage[nocompress]{cite}
\else
  \usepackage{cite}
\fi

\usepackage{tabularx}
\usepackage{multirow}
\usepackage{multicol}

\usepackage{enumitem}

\usepackage{subfig}
\usepackage{graphicx}
\usepackage{amsmath}
\usepackage{amssymb}
\usepackage{xcolor}
\usepackage{booktabs}

\usepackage{tabularx}

\usepackage[normalem]{ulem}
\useunder{\uline}{\ul}{}
\usepackage{url}  
\usepackage{xspace}
\makeatletter
\DeclareRobustCommand\onedot{\futurelet\@let@token\@onedot}
\def\@onedot{\ifx\@let@token.\else.\null\fi\xspace}
\def\eg{\emph{e.g}\onedot} 
\def\ie{\emph{i.e}\onedot}

\def\etal{\emph{et al}\onedot}
\makeatother

\DeclareMathOperator{\atantwo}{atan2}
\DeclareMathOperator{\sign}{sign}

\usepackage{pifont}

\newcommand{\PreserveBackslash}[1]{\let \temp =\\#1 \let \\ = \temp}
\newcolumntype{C}[1]{>{\PreserveBackslash\centering}p{#1}}
\newcolumntype{R}[1]{>{\PreserveBackslash\raggedleft}p{#1}}
\newcolumntype{L}[1]{>{\PreserveBackslash\raggedright}p{#1}}

\usepackage{diagbox}

\ifCLASSINFOpdf
\else
\fi
\usepackage{algorithmic}
\begin{document}
\title{Geometry-Guided Street-View Panorama Synthesis from Satellite Imagery}

\author{Yujiao~Shi,
        Dylan~Campbell,
        Xin~Yu,
        and~Hongdong~Li
% \author{Michael~Shell,~\IEEEmembership{Member,~IEEE,}
%         John~Doe,~\IEEEmembership{Fellow,~OSA,}
%         and~Jane~Doe,~\IEEEmembership{Life~Fellow,~IEEE}% <-this % stops a space
\IEEEcompsocitemizethanks{\IEEEcompsocthanksitem 
Y. Shi and H. Li are with the Australian National University. \protect\\%
E-mail: \{firstname.lastname\}@anu.edu.au.
\IEEEcompsocthanksitem D. Campbell is with the University of Oxford.\protect\\ E-mail: dylan@robots.ox.ac.uk.
\IEEEcompsocthanksitem X. Yu is with the University of Technology Sydney.\protect\\
 E-mail: xin.yu@uts.edu.au
}% <-this % stops an unwanted space
% \thanks{Manuscript received April 19, 2005; revised August 26, 2015.}
}


\markboth{ }%
{ }
\IEEEtitleabstractindextext{%
\begin{abstract}
This paper presents a new approach for synthesizing a novel street-view panorama given a satellite image, as if captured from the geographical location at the center of the satellite image.
Existing works approach this as an image generation problem, adopting generative adversarial networks to implicitly learn the cross-view transformations, but ignore the geometric constraints.
In this paper, we make the geometric correspondences between the satellite and street-view images explicit so as to facilitate the transfer of information between domains.
Specifically, we observe that when a 3D point is visible in both views, and the height of the point relative to the camera is known, there is a deterministic mapping between the projected points in the images.
Motivated by this, we develop a novel satellite to street-view projection (S2SP) module which learns the height map and projects the satellite image to the ground-level viewpoint, explicitly connecting corresponding pixels.
With these projected satellite images as input, we next employ a generator to synthesize realistic street-view panoramas that are geometrically consistent with the satellite images.
Our S2SP module is differentiable and the whole framework is trained in an end-to-end manner.
Extensive experimental results on two cross-view benchmark datasets demonstrate that our method generates more accurate and consistent images than existing approaches.
\end{abstract}

}

\maketitle

\IEEEdisplaynontitleabstractindextext

% \fi
\IEEEpeerreviewmaketitle

\IEEEraisesectionheading{\section{Introduction}\label{sec:introduction}}

\begin{figure*}
    \centering
    \subfloat[ \small Satellite \protect\\ \centering{Image}]{
    \centering
    \begin{minipage}{0.1\linewidth}
    \parbox[][2.2cm][c]{\linewidth}{
    \includegraphics[width=\linewidth]{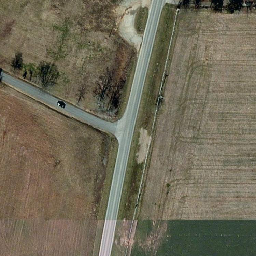}
    }
    \parbox[][2.2cm][c]{\linewidth}{
    \includegraphics[width=\linewidth]{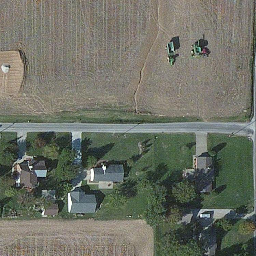}
    }
    \end{minipage}
    \label{satellite}
    }
    \subfloat[\small{\color{black}Height} \protect\\ \centering{{\color{black}Map}}]{
    \centering
    \begin{minipage}{0.1\linewidth}
    \parbox[][2.2cm][c]{\linewidth}{
    \includegraphics[width=\linewidth]{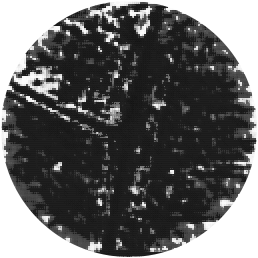}
    }
    \parbox[][2.2cm][c]{\linewidth}{
    \includegraphics[width=\linewidth]{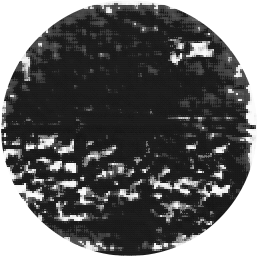}
    }
    \end{minipage}
    \label{intro_height_map}
    }
    \hfil
    \subfloat[ \small  Projected Satellite Image \protect\\ \centering{ (Intermediate)}]{
    \centering
    \begin{minipage}{0.245\linewidth}
    \parbox[][2.2cm][c]{\linewidth}{
    \includegraphics[width=\linewidth, height=0.4\linewidth]{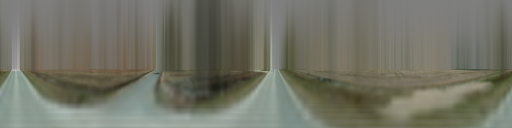}
    }
    \parbox[][2.2cm][c]{\linewidth}{
    \includegraphics[width=\linewidth, height=0.4\linewidth]{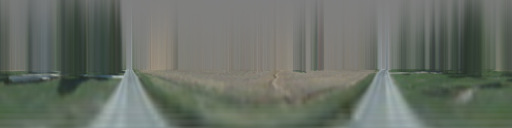}
    }
    \end{minipage}
    \label{Projected satellite}
    }
    \hfil
    \subfloat[ \small  Generated Street-View Image \protect\\ \centering{(Final)}]{
    \centering
    \begin{minipage}{0.245\linewidth}
    \parbox[][2.2cm][c]{\linewidth}{
    \includegraphics[width=\linewidth, height=0.4\linewidth]{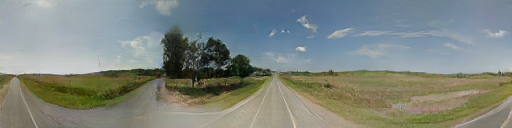}
    }
    \parbox[][2.2cm][c]{\linewidth}{
    \includegraphics[width=\linewidth, height=0.4\linewidth]{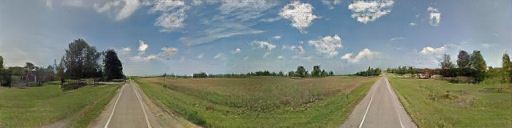}
    }
    \end{minipage}
    \label{Generated ground}
    }
    \hfil
    \subfloat[ \small  Real Street-View Image \protect\\ \centering{(Ground Truth)}]{
    \centering
    \begin{minipage}{0.245\linewidth}
    \parbox[][2.2cm][c]{\linewidth}{
    \includegraphics[width=\linewidth, height=0.4\linewidth]{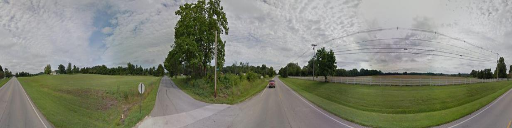}
    }
    \parbox[][2.2cm][c]{\linewidth}{
    \includegraphics[width=\linewidth, height=0.4\linewidth]{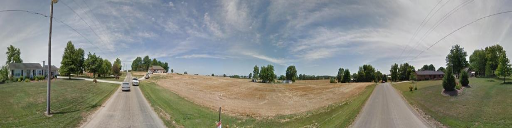}
    }
    \end{minipage}
    \label{Real ground}
    }
    \caption{\small Given a satellite image (a), our method first estimates the height distribution (b), where lighter is higher, and then differentiably projects the satellite image to the ground-level viewpoint (c) according to the estimated height distribution. Conditioned on the projected image, our generator synthesizes a realistic street-view panorama (d) that is geometrically consistent with the satellite image, and very similar to the real street-view image (e).
    }
    \label{fig:intro}
\end{figure*}

\IEEEPARstart{G}{iven} a satellite image, such as Figure~\ref{satellite}, what would one see when standing at the location of the image center?
In this example, one would reason that there is a fork in the road, a tree inside the fork, and grass elsewhere. 
This satellite to street-view image synthesis task aims to generate an omni-directional street-view panorama captured at a location corresponding to the center of the given satellite image. 
Our goal in this work is to synthesize a street-view panorama with scene structures that are as geometrically consistent with the satellite image as possible, while preserving visual similarity with the ground-truth panorama.

{
\color{black}
Structure-preserving street-view panorama synthesis is useful for several downstream tasks.
For instance, it has been shown to be helpful for cross-view image localization~\cite{Regmi_2019_ICCV, toker2021coming}.
For example, using both synthesized street-view images and the original satellite images can help to increase geo-localization performance~\cite{Regmi_2019_ICCV}. 
Combining satellite-to-street-view image synthesis with cross-view geo-localization in a unified framework further improves the performance of both tasks~\cite{toker2021coming}. 
Furthermore, satellite imagery now covers the entire world and is easily accessible almost everywhere. 
In contrast, street-view images are expensive to collect and are not available everywhere. 
Synthesizing ground-level images from satellite images helps to enrich media content for regions that are hard or expensive for humans or vehicles to access.
}

As a special case of novel view synthesis (NVS), the satellite to street-view image synthesis is remarkably challenging because:
(1) the significant change in viewing angle results in minimal field-of-view (FoV) overlap; and
(2) there are radical differences in image appearance since the imaging modalities are highly distinct and 
they may be captured at different times of day, seasons, and weather conditions. 

Conventional methods for solving this problem are often based on conditional generative adversarial networks (GANs) and they approach this problem as a pure image generation task. 
In their implementations, a powerful generator is employed to map the satellite images to the ground-level viewpoint by playing a min-max game with a discriminator. 

Although a deep neural network is theoretically able to learn any transformation,
neglecting the significant domain discrepancy between satellite and street-view images would lead to inferior performance.
To be specific, satellite images show the top of objects in an overhead view by parallel projection, while street-view panoramas capture scenes at ground level with a spherical equirectangular projection.
The transformations between satellite and street-view images are far more complex: they are not only content-dependent but also geometry-dependent.

In this paper, we develop a novel approach to explicitly establish the geometric correspondences between satellite and street-view images, recover the spatial layout information of scene objects in street-view, and handle occlusions. 
We achieve this goal by developing a satellite to street-view projection (S2SP) module. 
Our S2SP module first estimates height maps of the satellite images and uses these to project satellite image pixels to the street view. 

Specifically, our S2SP module estimates the height probability distribution for a given satellite image at a fixed set of heights, and constructs a satellite-view multiplane image (MPI) across the discretized heights to model the 3D scene. 
To generate the street-view panorama, the S2SP module then transforms the satellite-view MPI to a street-view MPI by unrolling and stretching a set of concentric cylinders. The street-view image is then rendered from the street-view MPI by employing an \emph{over} alpha compositing technique~\cite{porter1984compositing} in a back-to-front order. 
In this manner, our approach handles occlusions between scene objects in a differentiable way.
Figures~\ref{intro_height_map} and \ref{Projected satellite} provide two examples of the estimated height maps and the projected images from our S2SP module.
Conditioned on the projected images, we use a generator network to synthesize realistic panoramic images and in-paint the missing textures.

Furthermore, due to the issue of GPS drift, it is difficult to collect location-aligned satellite and street-view image pairs, where the location of the street-view camera exactly corresponds to the center of a satellite image \cite{lu2020geometry}. In this paper, our S2SP module also provides a method to align the collected satellite and street-view image pairs, providing clean training signals and accurate evaluation measurements for satellite to street-view synthesis.

This is an original submission that is not based on any published conference papers. 
The main contributions of this paper are summarized as follows:
\begin{itemize}[nosep, leftmargin=*]%, leftmargin=*
\item a new geometry-aware framework for satellite to street-view image synthesis, which explicitly establishes the geometric correspondences between the satellite and street-view images and allows a generator to focus on learning scene content dependent transformations (\ie, the visual appearance transformation of scene objects between the overhead view and the street view);
\item a novel and differentiable satellite to street-view image projection module, which provides a way to estimate the height map for satellite images and the visibility of objects in the street view, without explicit supervision; and
\item a novel mechanism for alleviating cross-view image pair misalignments, which not only helps to obtain clean satellite to street-view image synthesis training pairs but also identifies the location shift between the street-view camera and the satellite image center.
\end{itemize}

\begin{figure}[!t]
\setlength{\abovecaptionskip}{0pt}
\setlength{\belowcaptionskip}{0pt}
    \centering
    \includegraphics[width=\linewidth]{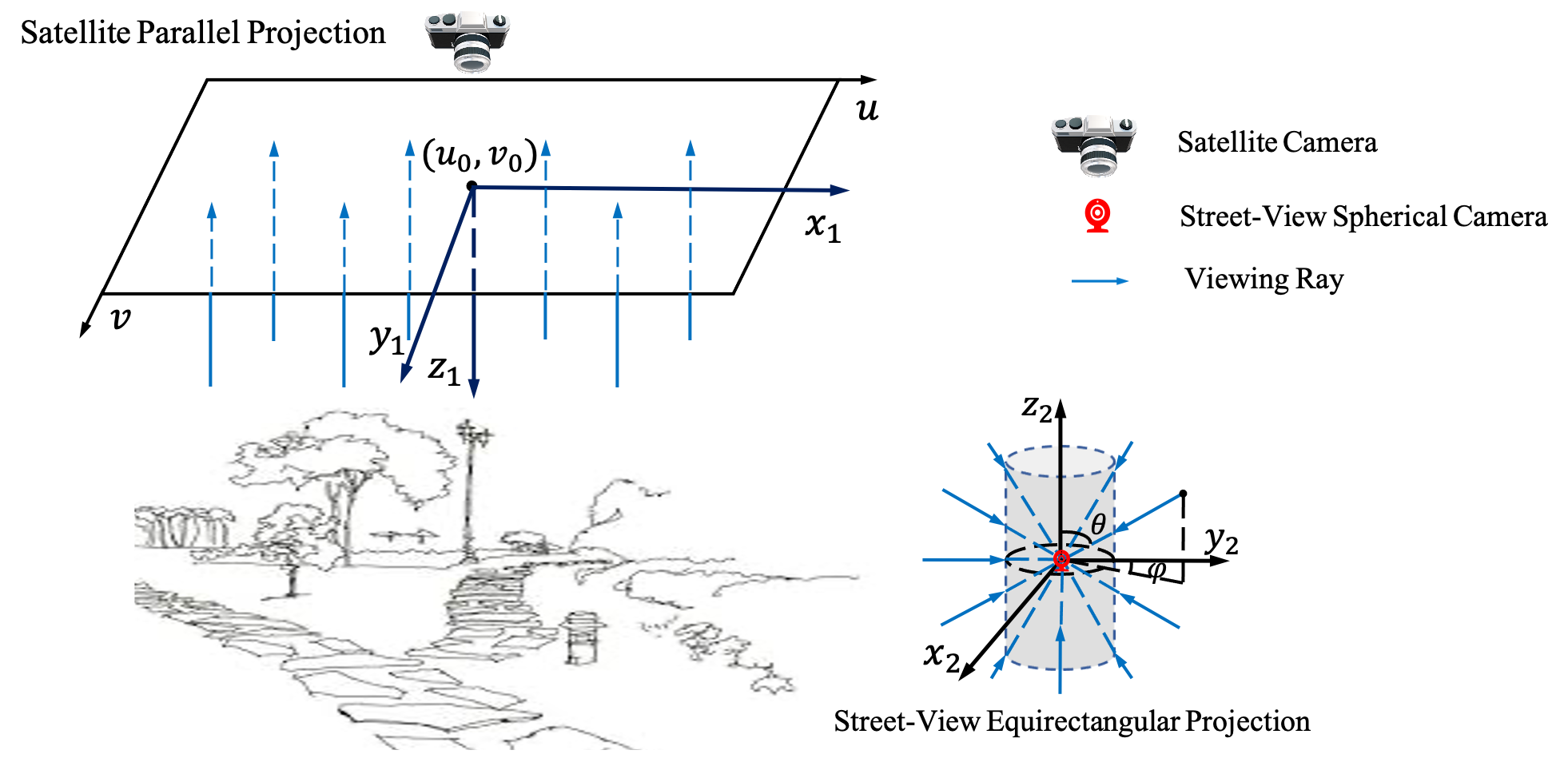}
    \caption{\small Visualization of different projection approaches by a satellite camera and a street-view spherical camera. The former is a parallel projection in the overhead view, and the latter is an equirectangular projection at ground level.}
    \label{fig: Sate-parallel and street-omnidirection.}
\end{figure}

\section{Related Work}
\subsection{Novel view synthesis}
Traditional novel view synthesis addresses the problem where only a small camera movement exists between source and target views. 
Liu \etal~\cite{liu2018geometry} tackled the problem of single image novel view synthesis. They approximated the real world scene as a set of planes with different surface normals, and learned to predict the homography of these planes, with which the input view can be transformed to the target view. 
Zhou \etal~\cite{zhou2018stereo} proposed the multiplane image representation (MPI) to extrapolate views from stereo images with narrow baselines. They modeled the scene as a number of image planes at a fixed range of depths with respect to a reference camera coordinate frame. 
Based on this representation, Flynn \etal~\cite{flynn2019deepview} explored a learned gradient method to estimate an MPI from a set of sparse camera viewpoints, and
Tucker and Snavely~\cite{tucker2020single} introduced a scale-invariant view synthesis mechanism to generate an MPI from a single-view online video.

Our method for satellite to street-view image synthesis is also built upon the multiplane image representation. 
As the viewpoint change is very large in our task, it is very difficult to directly render the target street-view panorama from an overhead-view MPI. 
Therefore, we integrate the overhead-view MPI as an intermediate output of our network rather than the final output, and employ a subsequent generator conditioning on the projected street-view images to inpaint missing textures and synthesize realistic images.

\subsection{Satellite-view and street-view synthesis}
The cross-view image synthesis task is extremely challenging between overhead and ground views since the visual appearance changes significantly.
Zhai \etal~\cite{zhai2017predicting} proposed to learn a linear transformation matrix between satellite and street-view semantics so that one can predict the street-view semantic layout by a matrix multiplication of the transformation and the satellite semantics.
Regmi and Borji~\cite{regmi2018cross} investigated employing conditional GANs for cross-view synthesis. Instead of a single image, their networks
regressed the target view image and its semantics jointly, with the semantic branch providing an additional supervision signal for image synthesis. 
{\color{black}{They further extended their work by using a homography to map the images based on the common area between the views~\cite{regmi2019crossview}. This provided more realistic details to the input image of conditional GANs. In contrast to a single homography on the ground plane, our method models geometric correspondences for all pixels between the views.} }
Tang \etal~\cite{tang2019multi} used street-view semantics and the satellite image to synthesize a street-view image.
{\color{black} 
In contrast, our approach does not require a target view semantic segmentation during training and testing, thus reducing the requirements of the dataset annotation (or a dataset used to pre-train a semantic segmentation network).
}
Instead, we exploit the two-view geometric constraints as a source of information with which to condition the generator. 
Lu \etal~\cite{lu2020geometry} also exploited geometric cues for satellite to street-view image synthesis, but they required ground-truth height and semantic supervision for the satellite images, and cannot train their network in an end-to-end manner due to discretized operations in the satellite-view to street-view projection. 
On the contrary, our method solves the problem in an end-to-end fashion under a more challenging and general setting, where the satellite map height supervision is not available.

\subsection{Cross-view image geo-localization}
Satellite and street-view image pairs are also frequently used for the task of image geo-localization.
Different from satellite to street-view image synthesis, the cross-view image geo-localization is a deep metric learning problem which aims to learn discriminative feature representations for scenes at different locations.
In this task, a query image captured by a ground-level camera is matched against a database with geo-tagged satellite images to determine the ground camera's location.

Workman and Jacobs~\cite{workman2015wide, workman2015location}
and Vo and Hays~\cite{vo2016localizing} pioneered this task by investigating a family of deep learning methods for the cross-view geo-localization. Hu \etal~\cite{Hu_2018_CVPR}, Sun \etal~ \cite{sun2019geocapsnet} and Cai \etal~\cite{Cai_2019_ICCV} focused on designing powerful networks or losses to achieve better results. Based on a simple deep architecture, Liu and Li~\cite{Liu_2019_CVPR} used orientation information of street-view and satellite images to assist geo-localization. 
Regmi and Shah \cite{Regmi_2019_ICCV} employed a conditional GAN to generate satellite images from street-view panoramas so as to bridge the domain gap for cross-view image matching.

Our previous works are based on cross-view image geo-localization, a different but related task to satellite to street-view image synthesis. 
For the cross-view image geo-localization, we first proposed an optimal feature transport module that aligned cross domain features to facilitate similarity matching~\cite{shi2020optimal}, and later showed that a polar transform could be used to establish approximate geometric correspondences between satellite-view and street-view images and thus boost the localization performance~\cite{shi2019spatial, shi2020looking}. 
The polar transform is a simple approximation of the nonlinear transformation between satellite-view and street-view images; in this work we instead use the true transformation from two-view geometry, which does not distort the scene geometry, for the image synthesis task.

\section{Satellite and Street-View Geometry}
\label{geometry}
\subsection{Parallel projection of a satellite camera}
As shown in Figure~\ref{fig: Sate-parallel and street-omnidirection.} (left), 
we denote the satellite camera coordinates as $(x_1, y_1, z_1)$ and the satellite image coordinates as $(u, v)$. 
The projection between the satellite camera coordinate system and satellite image coordinate system is approximated as a parallel projection, which maps the point $(x_1, y_1, z_1)$ to the satellite image point $(u, v)$ as

\begin{equation}
 \label{world2aer}
 \begin{split}
     u &= u_0 + sx_1 \\
     v &= v_0 + sy_1,
 \end{split}
\end{equation}
where $(u_0, v_0)$ is the satellite image center, and $s$ is the scale factor between the satellite image coordinates and the world coordinates.

\subsection{Perspective projection of an omnidirectional street-view camera}

The projection method of an omnidirectional street-view camera is illustrated in Figure~\ref{fig: Sate-parallel and street-omnidirection.} (right).
{\color{black} 
We use a cylinder to represent the image plane, but the pixels are parameterized under a spherical coordinate system.}
Let $(x_2, y_2, z_2)$ be the camera coordinates, and $(\theta, \phi)$ be the street-view image coordinates.
The omnidirectional street-view camera maps the point $(x_2, y_2, z_2)$ to the panorama image point $(\theta, \phi)$ by an equirectangular projection
\begin{equation}
 \label{world2grd}
 \begin{split}
      \theta &= 
    \left\{
\begin{array}{ll}
\atantwo ( \sqrt{x_2^{2} + y_2^{2}} , z_2)  &\hfil z_2 \neq 0 \\
\pi/2 &\hfil z_2=0 \\
\end{array}
\right . 
\\
\phi &= 
\left\{
\begin{array}{ll}
\atantwo (x_2, y_2) &\hfil y_2\neq 0 \\
\pi/2 \cdot \sign(x_2) &\hfil y_2=0 \\
\end{array}
\right . .
 \end{split}
\end{equation}

\subsection{Satellite-view and street-view geometry}

\begin{figure}[t]
\setlength{\abovecaptionskip}{0pt}
\setlength{\belowcaptionskip}{0pt}
    \centering
    \includegraphics[width=0.9\linewidth]{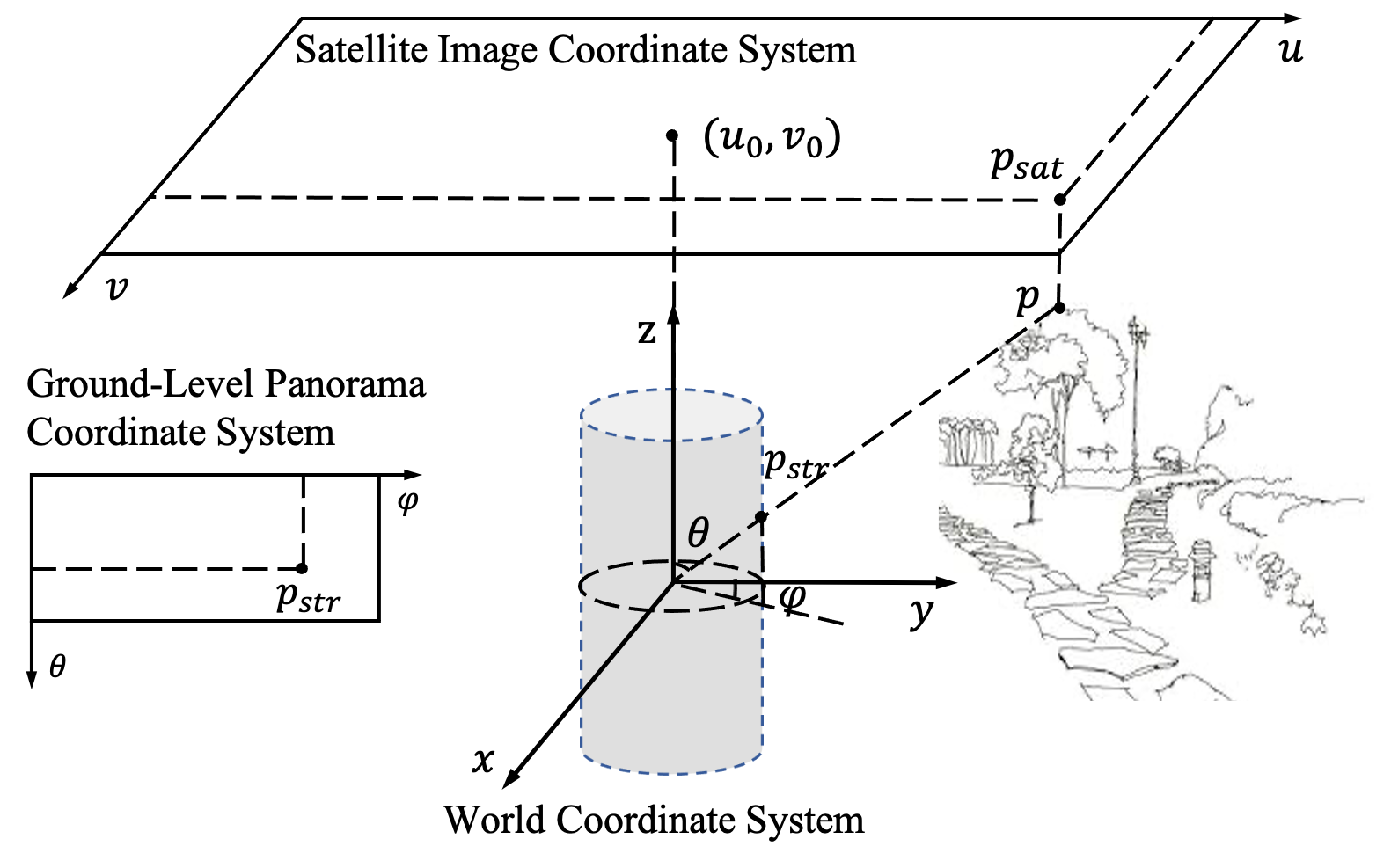}
    \caption{\small Corresponding pixels in a pair of satellite and street-view images.
    When a point $p$ in the world coordinate is visible in both the satellite and street-view images, there is a deterministic geometric mapping between the projected pixels $p_\text{sat}$ and $p_\text{str}$.
    }
    \label{fig:geometric correspondences}
\end{figure}

For the satellite to street-view image synthesis, we illustrate the geometric correspondences between satellite and street-view images in Figure~\ref{fig:geometric correspondences}. 
As shown in the figure, the street-view camera is set at the location corresponding to the satellite image center.
We set the origin of the world coordinate to the ground camera location, where its $x$-axis is parallel to the $v$ direction of the satellite image coordinate, $y$-axis is parallel to the $u$ direction of the satellite image coordinate, and $z$ axis points upward.
For a pixel $p_\text{sat}$ in the satellite image which is visible from the street view, the transformation between $p_\text{sat}$ and its projected location at the street-view panorama $p_\text{str}$ is deterministic given the height $z$, expressed as
\begin{equation}
 \label{sat2str}
 \begin{split}
      \theta &= 
    \left\{
\begin{array}{ll}
\atantwo ( \sqrt{(v-v_0)^{2} + (u-u_0)^{2}} , sz)  &\hfil z \neq 0 \\
\pi/2 &\hfil z=0 \\
\end{array}
\right . 
\\
\phi &= 
\left\{
\begin{array}{ll}
\atantwo (v-v_0, u-u_0) &\hfil u\neq u_0 \\
\pi/2 \cdot \sign(v-v_0) &\hfil u=u_0 \\
\end{array}
\right . .
 \end{split}
\end{equation}

\begin{figure*}[!t]
\setlength{\abovecaptionskip}{0pt}
    \setlength{\belowcaptionskip}{0pt}
    \centering
    \includegraphics[width=0.9\linewidth]{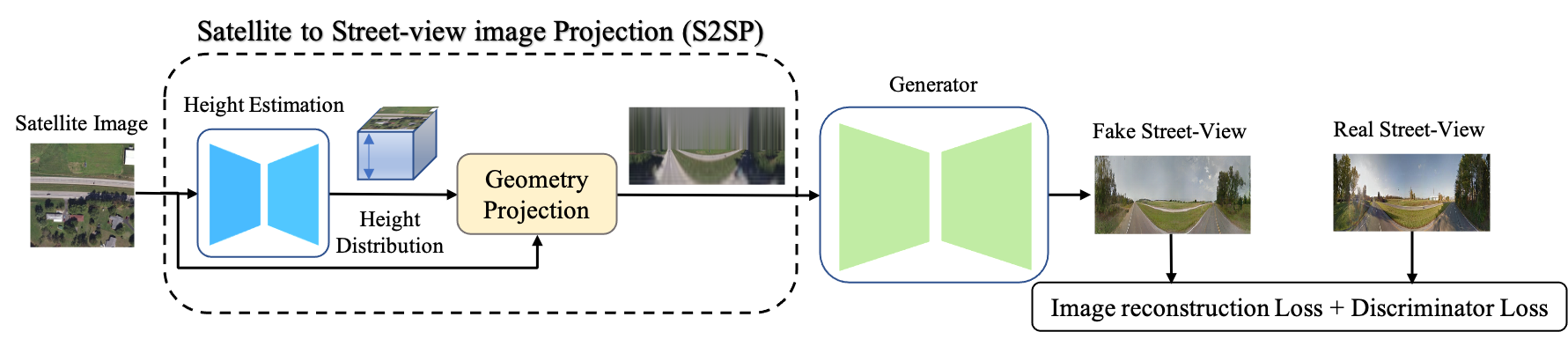}
    \caption{\small Flowchart of the proposed framework. We propose a novel satellite to street-view image projection module to transform the satellite images to the street viewpoint and a projection-conditioned generator to synthesize realistic street-view panoramas that are geometrically consistent with the satellite images.}
    \label{fig:framework}
\end{figure*}

In order to establish these geometric correspondences, one could estimate a pixel-wise height map for the satellite image and then project the corresponding 3D points into the street-view image plane.
A standard approach would be to sort the projected 3D points along each viewing ray by depth (using a z-buffer) and selecting the closest point to color the corresponding pixel.
However, this forward mapping approach has several problems.
(1) Poor occlusion modeling: this approach only models the top surface of the scene, and so the occlusion in the global world is hard to be modeled.
For example, points that lie behind a tree (viewing at the ground level) will not be occluded, because only the tree canopy is treated as an occlusion surface.
(2) Forward mapping artifacts: the significant mismatch in resolution between the satellite and street-view images leads to many missing pixels in the output projected image.
(3) Closest point selection: hard selection from the z-buffer is non-differentiable near the visibility boundary, because a small change in the estimated height can lead to a discontinuous change in the output projection, as points shift from visible to invisible and vice versa. Moreover, it leads to sparse gradient signals, since there is zero gradient for non-visible points. This is sub-optimal for points that should be visible but are not, due to a slightly incorrect height estimate.

Considering these issues, we instead propose to model the scene with a dense volumetric representation rather than sparse 3D points, and use an inverse mapping to obtain the street-view projection.
We use multiplane images (MPIs)~\cite{zhou2018stereo} to achieve this goal.

An MPI is a set of fronto-parallel image planes $\{I_1, I_2, ..., I_N\}$ at a fixed range of depths with respect to a reference coordinate frame, where $N$ is the number of the depth planes. 
Each image plane $I_i$ in an MPI includes three color channels $C_i\in \mathbb{R}^{H \times W \times 3}$ and an alpha channel $\alpha_i \in \mathbb{R}^{H \times W \times 1}$ to encode the transparency, where $H$ and $W$ are the height and width of the image planes.
This multiplane image structure is able to represent the geometry and texture of scenes including occluded elements.
In the next section, we will present the technical details on the MPI construction and its usage in our proposed differentiable satellite to street-view image projection.

\section{Street-View Synthesis Guided By Two-View Geometry}
\label{method}

An overview of our network is shown in Figure~\ref{fig:framework}.
Our method establishes the geometric correspondences between the satellite and street-view images by a differentiable Satellite to Street-view image Projection (S2SP) module, and then employs a generator to produce realistic and geometrically consistent street-view panoramas with respect to input satellite images.

\subsection{Satellite to street-view image projection}

\textbf{Height estimation.}
Given a satellite image, we first estimate its pixel-wise height probability distribution for a fixed range of heights, given by
\begin{equation}
\begin{array}{lr}
    D = f_\text{height}(I_\text{sat}) & \text{s.t. } 
    \sum_{i=1}^{N}D_{h,w,i}=1,
\end{array}
\end{equation}
where $D \in \mathbb{R}^{H^{\color{black}\text{sat}} \times W^{\color{black}\text{sat}} \times N}$ denotes the estimated probability distribution of the $N$ discretized heights, 
$f_\text{height}(\cdot)$ is the height estimation network, 
$ I_\text{sat}\in \mathbb{R}^{H^{\color{black}\text{sat}} \times W^{\color{black}\text{sat}} \times 3}$
denotes the RGB satellite image, and 
$h$, $w$ and $i$ 
are the indices for the 
image height, image width and height (elevation) 
dimensions, respectively.
Estimating the height probability of pixels instead of a single height map enables us to model the 3D scene in a dense and relatively continuous manner. 

\begin{figure*}[ht]
\setlength{\abovecaptionskip}{0pt}
\setlength{\belowcaptionskip}{0pt}
    \centering
    \includegraphics[width=0.9\linewidth]{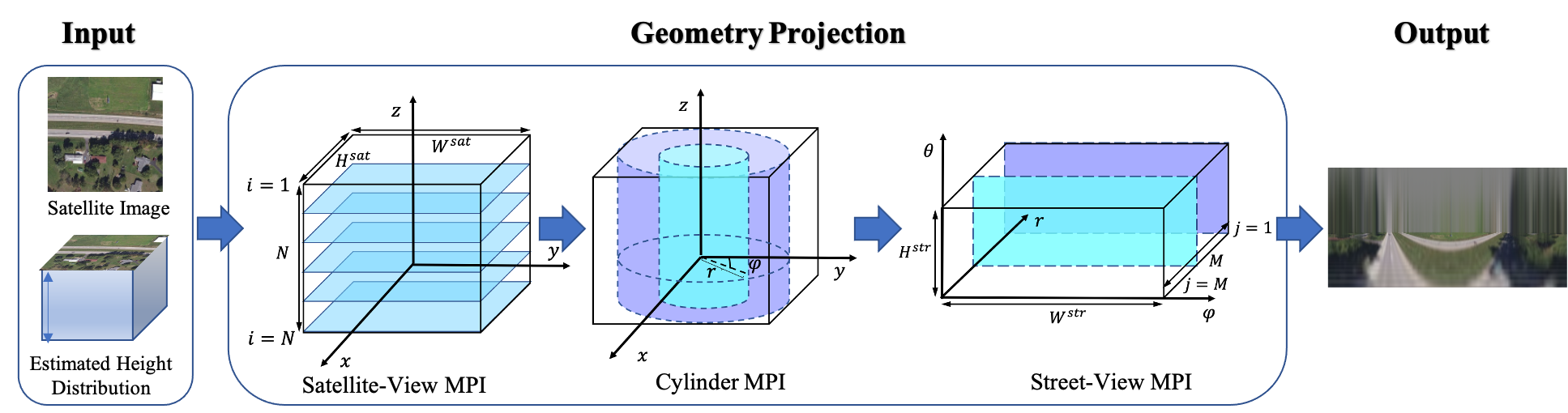}
    \caption{\small Illustration of the geometry projection block in our S2SP module. We first construct an overhead-view MPI according to the given satellite image and the estimated height probability distribution, and then convert it to the street viewpoint by unrolling and stretching each concentric cylinder. The final street-view image is rendered in a back-to-front order from the street-view MPI.}
    \label{fig:geometric_transfer}
\end{figure*}

\smallskip
\noindent\textbf{Satellite-view MPI construction.}
Given the estimated height probability distribution, we construct a satellite-view MPI across the discretized heights to model the global scene. 
As illustrated in Figure~\ref{fig:geometric_transfer}, an satellite-view MPI is composed of a set of image planes parallel to the satellite image with different heights. 

Pixels in a satellite image correspond to the topmost points of scene objects. 
Purely modeling the 3D points corresponding to the satellite image pixels is insufficient for solving the occlusion problem at the street view, 
{\color{black} since it leaves the points below the top structures undetermined.}
Therefore, we also model the lower points by assuming that these points are also opaque.
% make an assumption that the 3D points directly below are also opaque.
This is achieved by setting the transparency channel for each image plane $\alpha_{i}^{{\color{black}\text{sat}}} \in \mathbb R ^{H^{\text{sat}}\times W^{\text{sat}}\times 1}$ to the cumulative height probability distribution
\begin{equation}
\alpha_{i}^{{\color{black}\text{sat}}} = \sum\nolimits_{k=1}^{i}D_{\cdot, \cdot, k},
\end{equation}
where plane $i=N$ is the ground plane.
For the color channels of the satellite-view MPI, we set them to the input satellite image, with $C_i^{{\color{black}\text{sat}}} = I_\text{sat}$.

\smallskip
\noindent\textbf{Street-view MPI projection.}
As our target is to synthesize street-view panoramas, our next step is to determine the order of these points along viewing rays at the street viewpoint.
To do so, we first decompose the overhead-view MPI to a set of concentric cylinders with uniformly sampled radii, as illustrated in Figure~\ref{fig:geometric_transfer} (middle), and then project the cylinders to the street-view panorama (spherical) image coordinates.
The result is a set of image planes at the street viewpoint with uniformly sampled depths (along the $z$-axis), which we refer to as the street-view MPI. 

Let $(u, v, z)$ denote the coordinates of points in the overhead-view MPI and $(\theta,\phi, r)$ represent the coordinates of projected points in the street-view MPI, where $r$ is the concentric cylinder radius and $r = \sqrt{(u-u_0)^2 + (v-v_0)^2}$. The transformation between the source points in the original satellite MPI and the target points in the street-view MPI is
\begin{equation}
 \label{aer2grdMPI}
 \begin{split}
     u &= u_0 + sr\cos \phi \\
     v &= v_0 + sr\sin \phi \\
     z &= 
        \left\{
        \begin{array}{ll}
        r/\tan \theta  & \theta \in [0, \pi/2) \cup (\pi/2, \pi]\\
        0 & \theta = \pi/2 \\
        \end{array}
        \right . .
 \end{split}
\end{equation}
With this transformation, the points are sorted from far to near along viewing rays at the street view, and the satellite to street-view image projection is solved by an inverse mapping.

\noindent\textbf{Street-view image rendering.}
{\color{black} Denote $C_j^{\text{str}} \in \mathbb R ^{H^{\text{str}}\times W^{\text{str}}\times 3}$ and $\alpha_j^{\text{str}}\in \mathbb R ^{H^{\text{str}}\times W^{\text{str}}\times 1}$ as the color and alpha channels of the street view MPI, respectively, where $j\in [1, M]$ is the index of street-view image planes, $M$ is the plane number of the street-view MPI, $H^{\text{str}}$ and $W^{\text{str}}$ are the height and width of target street-view images, respectively.}
The street-view image is then rendered from the street-view MPI using \emph{over} alpha compositing~\cite{zhou2018stereo} in a back-to-front order
\begin{equation}
\label{alpha_composite}
    I_\text{comp}^{j} =
    \begin{cases}
    \alpha_1^{{\color{black}{\text{str}}}} \cdot C_1^{{\color{black}{\text{str}}}} & {\color{black}\text{if}\: j=1}\\ 
    \alpha_j^{{\color{black}{\text{str}}}} \cdot C_j^{{\color{black}{\text{str}}}} + (1-\alpha_j^{{\color{black}{\text{str}}}}) \cdot I_\text{comp}^{j-1} & {\color{black}\text{otherwise}}
    \end{cases},
\end{equation}
where 
{\color{black}{$j=1$}}
indicates the furthest image plane and 
{\color{black}{$j=M$}}
corresponds to the closest image plane.
The final composited image is thus $I_\text{comp}^{{\color{black}M}}$.
Points with higher $\alpha$ values will have higher weights in the composited image and thus dominate the color of a pixel.
By using this approach, all the pixels in the original satellite image contribute to the final rendered street-view image, and thus it allows dense gradients during training. 
Since all the operations in our S2SP module are differentiable, our network can be trained in an end-to-end manner.

\subsection{Network architecture}
As illustrated in Figure~\ref{fig:framework}, 
our network includes a Satellite to Street-view image Projection (S2SP) module and a generator to synthesize geometrically consistent and realistic street-view panoramas.
We employ the Pix2Pix~\cite{isola2017image} network as our generator backbone.
For the height estimation block in the S2SP module, we employ the same architecture as the Pix2Pix network but with the number of output channels in each layer reducing by $1/16$.
Since ground-truth height maps of the satellite images are not available, 
there is no explicit supervision for the height estimation block in our pipeline. Instead, implicit supervision is provided by enforcing the two-view geometric constraints and propagating the error signal back from the output of the generator.
In particular, if the estimated heights of pixels deviate from ground-truth values, the projected 3D points will be reordered when viewed at the ground level, changing the rendered images. Thus, the error between the final generated image and the real target image may be larger when the heights are estimated incorrectly.
Backpropagating this signal through the differentiable cross-view projection layer allows the height estimation network parameters to be updated without explicit supervision.

\subsection{Adversarial learning from corresponding satellite and street-view image pairs}

Following recent state-of-the-art approaches \cite{regmi2018cross, tang2019multi}, we use a generative adversarial network and adopt an adversarial training procedure. 
The generator $G$ is trained to map satellite images to their street-view counterparts by playing a min-max game with the discriminator network $D$.
The generative adversarial objective function, minimized by the generator and maximized by the discriminator, is given by
\begin{equation}
\label{Gan_loss}
\begin{aligned}
\mathcal L_{GAN} (G, D) &= \mathbb E_{I_\text{str} \sim p_{str}(I_\text{str}) } [ \log{D(I_\text{str})} ]\\
&+ \mathbb E_{I_\text{sat} \sim p_{sat}(I_\text{sat}) } [ \log{(1-D(G(I_\text{sat})))}],
\end{aligned}
\end{equation}
where $I_\text{sat}$ is the input satellite image, $I_\text{str}$ is the real street-view image, and $G(I_\text{sat})$ is the generated street-view image.

Since the database street images do not strictly correspond to their satellite counterparts, \eg, the satellite images are captured in summer while the street-view panoramas depict scenes in winter or autumn, 
we adopt the perceptual loss $\mathcal L_\text{per}$ \cite{zhou2018stereo} {\color{black} along with the $\mathcal L_1$ loss} between the real and generated street-view images to evaluate their feature similarity and their pixel-wise color similarity.
{\color{black} Overall, the image reconstruction loss is given by}
\begin{align}
\label{perceptual_loss}
   \mathcal L_\text{rec}(G) &= \mathcal L_1(G) + \mathcal L_\text{per} (G)\\
 &=\| I_\text{str} - G(I_\text{sat}) \|_1  +\! \sum \nolimits_{l} \lambda_l \| d_l(I_\text{str}) \!- d_l(G(I_\text{sat})) \|_1,\nonumber
 \end{align}
where ${d_l(\cdot)}$ indicates a set of feature representations of an image from a VGG-19 network~\cite{Simonyan2014VeryDC} (conv1\_1, conv2\_2, conv3\_2, conv4\_2 and conv5\_2), $\| \cdot \|_1$ is the $L_1$ distance and ${\lambda_l}$ represents the 
corresponding weight hyperparameters. 
Following the work of Zhou \etal~\cite{zhou2018stereo}, we set the weight hyperparameters to the inverse of the number of neurons in each layer.
The overall objective of our network, minimized by the generator and maximized by the discriminator, is
\begin{equation}
\label{loss}
    \mathcal L(G, D) = \mathcal L_\text{GAN} (G, D) + \mathcal L_\text{rec} (G).
\end{equation}

\begin{figure*}[!ht]
    \setlength{\abovecaptionskip}{0pt}
    \setlength{\belowcaptionskip}{0pt}
    \centering
    \subfloat[\small Satellite\protect\\ \centering Image]{
    \centering
    \begin{minipage}{0.11\linewidth}
    \includegraphics[width=\linewidth]{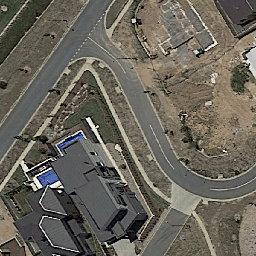}
    \end{minipage}
    \label{satellite original}
    }
    \hfil
    \subfloat[ \small Projected Satellite Image \protect\\ \centering (Original)]{
    \centering
    \begin{minipage}{0.275\linewidth}
    \includegraphics[width=\linewidth, height=0.4\linewidth]{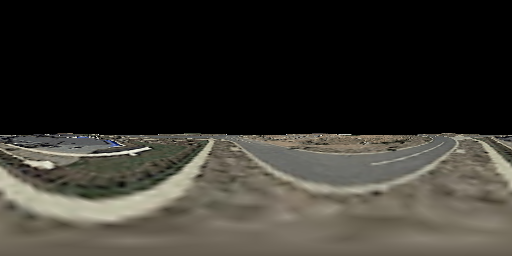}
    \end{minipage}
    \label{Projected satellite Original}
    }
    \hfil
    \subfloat[ \small Projected Satellite Image \protect\\ \centering (Corrected)]{
    \centering
    \begin{minipage}{0.275\linewidth}
    \includegraphics[width=\linewidth, height=0.4\linewidth]{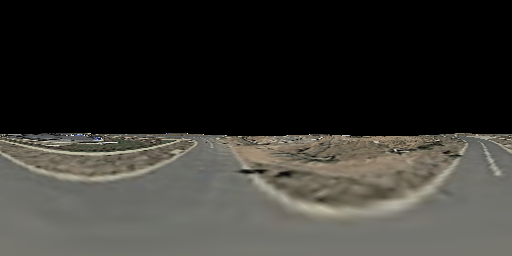}
    \end{minipage}
    \label{Projected satellite Rectified}
    }
    \hfil
    \subfloat[ \small Street-View Image]{
    \centering
    \begin{minipage}{0.275\linewidth}
    \includegraphics[width=\linewidth, height=0.4\linewidth]{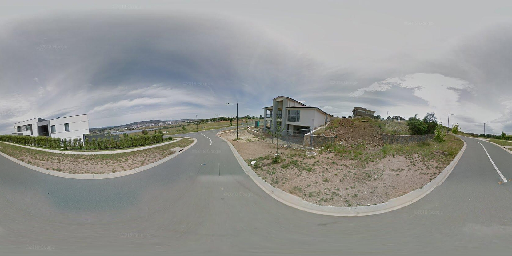}
    \end{minipage}
    \label{Street-View Image}
    }
    \caption{\small Example of satellite and street-view image misalignment and correction: (a) original satellite image, (b) projected satellite image according to the original satellite image center, (c) projected satellite image after mitigating the satellite and street-view image misalignment, and (d) corresponding street-view image in the database. The height maps are set to zero for this visualization.}
    \label{fig:alignment correction}
\end{figure*}

\begin{figure}[!t]
    \setlength{\abovecaptionskip}{0pt}
    \setlength{\belowcaptionskip}{0pt}
    \centering
    \subfloat[ \small Satellite Image]{
    \centering
    \begin{minipage}{0.28\linewidth}
    \includegraphics[width=\linewidth]{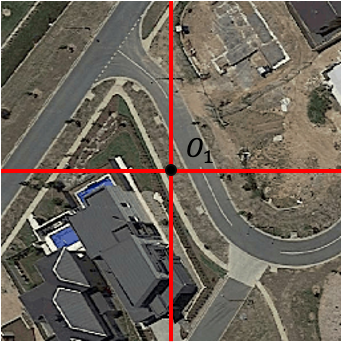}
    \end{minipage}
    \label{satellite original with centroid}
    }
    \hfil
    \subfloat[\small Projected \protect\\ \centering Street-View]{
    \centering
    \begin{minipage}{0.28\linewidth}
    \includegraphics[width=\linewidth, height=\linewidth]{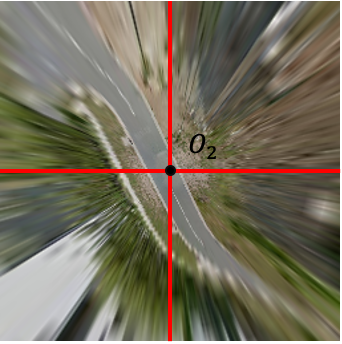}
    \end{minipage}
    \label{Projected ground}
    }
    \hfil
    \subfloat[\small Overlay]{
    \centering
    \begin{minipage}{0.28\linewidth}
    \includegraphics[width=\linewidth, height=\linewidth]{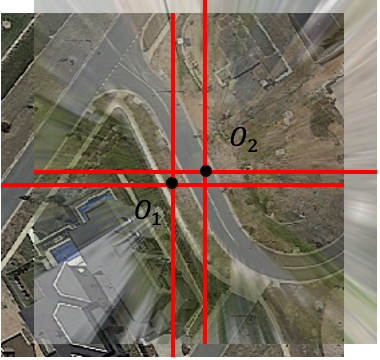}
    \end{minipage}
    \label{Overlay}
    }
    \caption{\small Visualization of satellite and street-view image pair misalignment: (a) original satellite image with image center $O_1$; (b) projected street-view image in the overhead view with image center $O_2$; and (c) overlaid image of (a) and (b). There is a location shift between $O_1$ and $O_2$.}
    \label{fig:alignment visualization}
\end{figure}

\section{Aligning Satellite and Street-View Image Pairs}
\label{mislignment fixing}
Due to the GPS positioning error, it is hard to collect strictly location aligned satellite and street-view image pairs for satellite to street-view image synthesis, where the camera location of street-view panorama corresponds exactly to the center of the satellite image.
Figure~\ref{satellite original} and Figure~\ref{Street-View Image} show an example satellite and street-view image pair from the CVACT dataset,
which is tagged as ``matching'' but is not strictly aligned. 

This misalignment is detected using the following procedure. We first project the street-view panorama to the satellite view by exploiting the geometric correspondences illustrated in Section~\ref{geometry}. In this process, we assume that all the pixels lie on the ground plane. The results are presented in Figure~\ref{Projected ground}.
We then overlay it on the original satellite image (Figure~\ref{satellite original with centroid}), as shown in Figure~\ref{Overlay}.
There is a shift between the original satellite image center $O_1$ and the projected street-view image center $O_2$.

To correct this misalignment automatically, we exploit the satellite to street-view image projection.
Since the misalignment is small, we select $40 \times 40$ points in a central region of the original satellite image, corresponding to $11.25 \times 11.25$ meters, to model the potential shifts between them.
We next exhaustively project the satellite image into the street viewpoint at each of the points, and compare the similarity (SSIM value) between the projected images and the original street-view image.
The one with the maximum similarity is selected and its corresponding ``shift'' is adopted to align the cross-view image pair.

We perform geometric satellite to street-view image projection under the assumption (for this part only) that all the pixels lie on the ground plane.
Hence there are no trainable parameters and we can use this directly as a pre-processing step to create clean satellite and street-view training pairs.
Figure~\ref{Projected satellite Original} and Figure~\ref{Projected satellite Rectified} show a comparison of the projected images before and after our correction. It can be seen that the projected image after correction is aligned with the corresponding street-view image from the dataset.

\section{Benchmarks for Satellite to Street-View Image Synthesis}
There are currently two large-scale and publicly available cross-view datasets, namely, CVACT~\cite{Liu_2019_CVPR} and CVUSA~\cite{zhai2017predicting}.
These two datasets has been widely used as benchmarks for cross-view image based geo-localization~\cite{Hu_2018_CVPR, Cai_2019_ICCV, Regmi_2019_ICCV, shi2020optimal, shi2019spatial, shi2020looking, Liu_2019_CVPR, zhai2017predicting, sun2019geocapsnet}.
In this paper, we instead employ them for satellite to street-view image synthesis.

Both CVACT and CVUSA contain $35,532$ satellite and street-view image pairs for training and $8,884$ image pairs for testing.
As both of the two datasets are introduced for image geo-localization, they are allowed to have slightly location shift between matching satellite and street-view image pairs.
However, in the satellite to street-view image synthesis task, the goal is to synthesize a street-view panorama as if it is captured at the same geographical location as the satellite image center.
Therefore, it is necessary to have exactly location aligned cross-view image pairs, 
{{especially for performance evaluation. }}

For CVACT, we propose a new split for the problem of satellite to street-view image synthesis, with $26,519$ training pairs and $6,288$ testing pairs, and augment the dataset with translation offsets to provide ground-truth alignment. This new split is necessary because many of the ostensible ground-truth image pairs are not aligned in the original dataset, that is, the translation offset between the centre of the satellite image and the camera position is unknown. This is highly problematic for the cross-view synthesis task.
To generate this split, we automatically computed the translation offsets for the image pairs, using the image alignment method proposed in Section~\ref{mislignment fixing}. However, we are unable to compute the offset in two situations: (1) where the translational misalignment is too large, typically more than $16$ meters, and (2) where there is severe cross-class occlusion in the vertical direction. The latter refers to situations where, for example, the satellite view can only see a tree canopy, while the street view sees the road surface underneath. For these cases, our alignment method, and human annotators, are unable to estimate the translation offset, and so these image pairs are filtered from the split.
The split and translation offsets will be made publicly available.

The street-view panoramas in CVUSA dataset were cropped at the top and bottom by Zhai~\etal~\cite{zhai2017predicting} to reduce the fraction of sky and ground. It is not clear whether they have been cropped uniformly across the dataset, so we cannot apply the same misalignment rectification method to the CVUSA dataset.
During training and testing, we approximate the street-view panoramas in the CVUSA dataset as having a 90-degree vertical field of view (FoV) with the central horizontal line corresponding to the horizon. The CVACT dataset contains panoramas with a 180-degree vertical FoV \cite{Liu_2019_CVPR}.
Note that the panoramic images in both datasets have a 360-degree horizontal FoV.

Lu \etal~\cite{lu2020geometry} also proposed a cross-view dataset with ground truth semantics and height maps for satellite images (which are not available in CVACT and CVUSA dataset). However, this dataset has not been released. 
Regmi and Shah~\cite{Regmi_2019_ICCV} also introduced the OP dataset for fine-grained cross-view geo-localization. 
{\color{black} However, this dataset has significant and uncorrelated position and orientation misalignments and is an order of magnitude smaller than CVACT and CVUSA, making it unsuitable for the image synthesis task.}
Therefore, we use the CVACT and the CVUSA datasets for evaluation.

\section{Experiments}

\begin{figure*}[ht!]
\setlength{\abovecaptionskip}{0pt}
    \setlength{\belowcaptionskip}{0pt}
\centering
\subfloat[ \small CVACT (Aligned)]{
    \begin{minipage}{0.11\linewidth}
    \begin{minipage}[t][\linewidth][t]{\linewidth}
    \includegraphics[width=\linewidth]{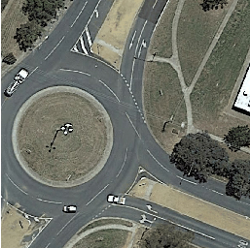}
    \end{minipage}
    \begin{minipage}[t][\linewidth][t]{\linewidth}
    \includegraphics[width=\linewidth]{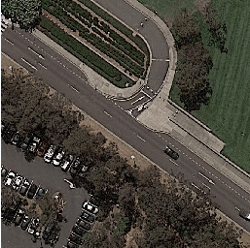}
    \end{minipage}
    \begin{minipage}[t][\linewidth][t]{\linewidth}
    \includegraphics[width=\linewidth]{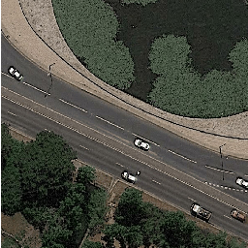}
    \end{minipage}
    \begin{minipage}[t][\linewidth][t]{\linewidth}
    \includegraphics[width=\linewidth]{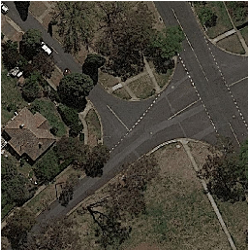}
    \end{minipage}
    \centerline{\small Satellite Image}
    \end{minipage}
    \begin{minipage}{0.22\linewidth}
    \includegraphics[width=\linewidth]{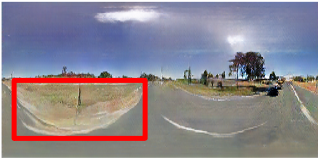}
    \includegraphics[width=\linewidth]{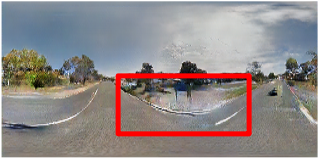}
    \includegraphics[width=\linewidth]{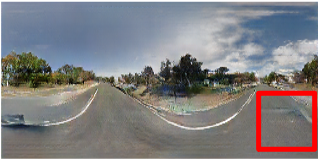}
    \includegraphics[width=\linewidth]{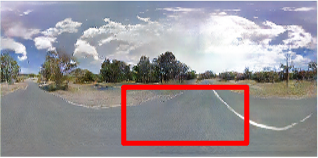}
    \centerline{\small Pix2Pix~\cite{isola2017image}}
    \end{minipage}
    \begin{minipage}{0.22\linewidth}
    \includegraphics[width=\linewidth]{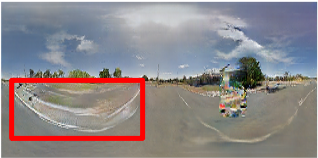}
    \includegraphics[width=\linewidth]{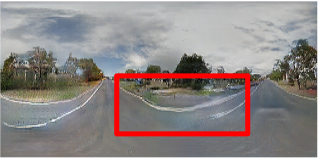}
    \includegraphics[width=\linewidth]{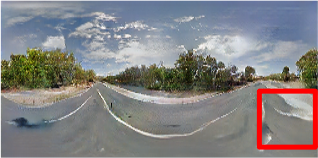}
    \includegraphics[width=\linewidth]{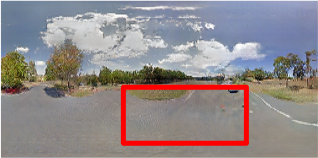}
    \centerline{\small XFork~\cite{regmi2018cross}}
    \end{minipage}
    \begin{minipage}{0.22\linewidth}
    \includegraphics[width=\linewidth]{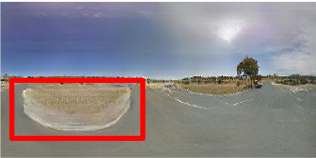}
    \includegraphics[width=\linewidth]{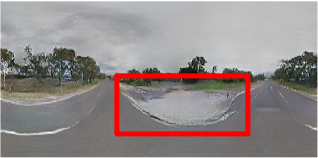}
    \includegraphics[width=\linewidth]{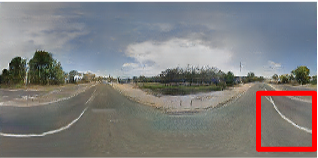}
    \includegraphics[width=\linewidth]{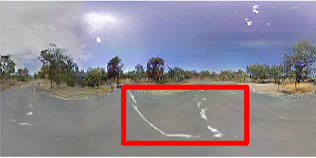}
    \centerline{\small Ours }
    \end{minipage}
    \begin{minipage}{0.22\linewidth}
    \includegraphics[width=\linewidth]{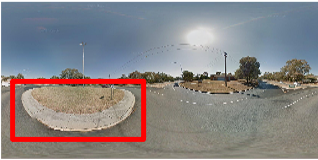}
    \includegraphics[width=\linewidth]{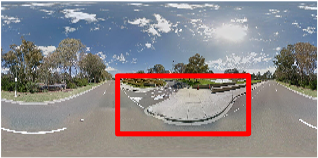}
    \includegraphics[width=\linewidth]{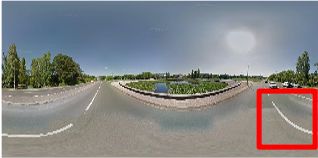}
    \includegraphics[width=\linewidth]{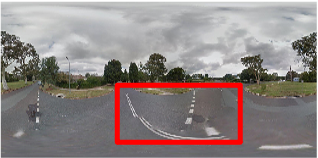}
    \centerline{\small Ground Truth}
    \end{minipage}
    \label{fig:visualization on CVACT}
    }\\
    \subfloat[ \small CVUSA]{
    \begin{minipage}{0.11\linewidth}
    \includegraphics[width=\linewidth]{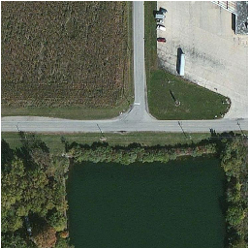}
    \includegraphics[width=\linewidth]{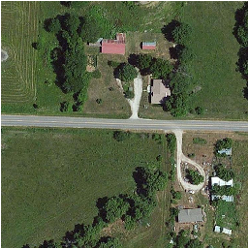}
    \includegraphics[width=\linewidth]{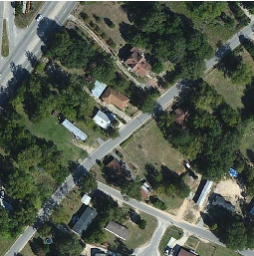}
    \includegraphics[width=\linewidth]{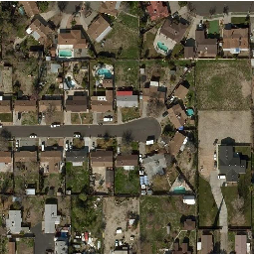}
    \centerline{\small Satellite Image}
    \end{minipage}
    \begin{minipage}{0.22\linewidth}
    \includegraphics[width=\linewidth]{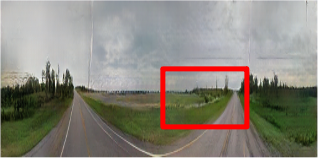}
    \includegraphics[width=\linewidth]{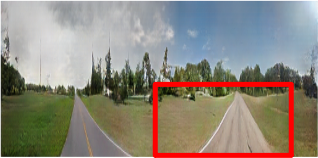}
    \includegraphics[width=\linewidth]{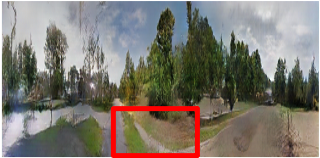}
    \includegraphics[width=\linewidth]{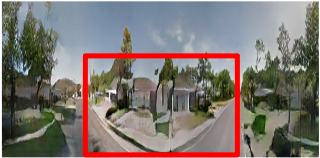}
    \centerline{\small Pix2Pix~\cite{isola2017image}}
    \end{minipage}
    \begin{minipage}{0.22\linewidth}
    \includegraphics[width=\linewidth]{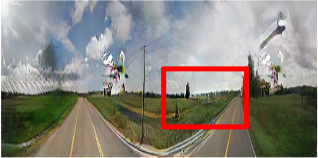}
    \includegraphics[width=\linewidth]{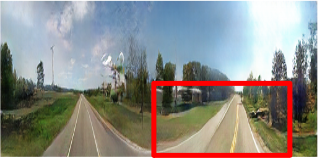}
    \includegraphics[width=\linewidth]{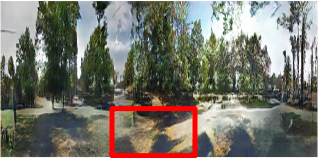}
    \includegraphics[width=\linewidth]{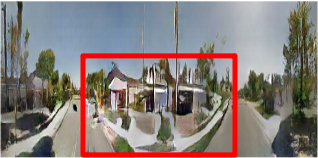}
    \centerline{\small XFork~\cite{regmi2018cross}}
    \end{minipage}
    \begin{minipage}{0.22\linewidth}
    \includegraphics[width=\linewidth]{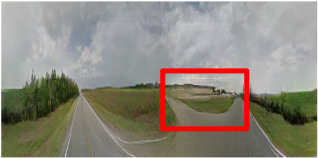}
    \includegraphics[width=\linewidth]{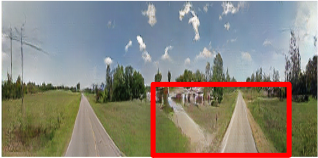}
    \includegraphics[width=\linewidth]{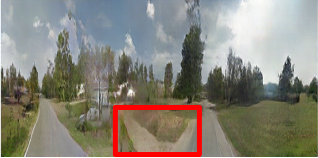}
    \includegraphics[width=\linewidth]{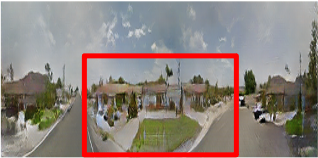}
    \centerline{\small Ours}
    \end{minipage}
    \begin{minipage}{0.22\linewidth}
    \includegraphics[width=\linewidth]{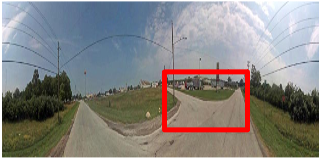}
    \includegraphics[width=\linewidth]{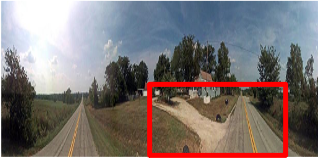}
    \includegraphics[width=\linewidth]{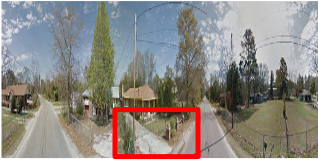}
    \includegraphics[width=\linewidth]{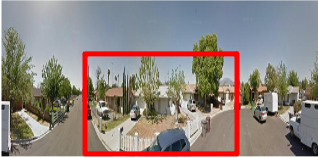}
    \centerline{\small Ground Truth}
    \end{minipage}
    \label{fig:visualization on CVUSA}
    }\\
    \caption{\small Qualitative comparison of synthesized images for the CVACT (Aligned) and CVUSA datasets.}
    \label{fig:Qualitative comparison}
\end{figure*}

\begin{table*}[ht!]
\setlength{\abovecaptionskip}{0pt}
\setlength{\belowcaptionskip}{0pt}
\centering
\caption{\small Quantitative comparison with existing algorithms on the CVACT (Aligned) and CVUSA datasets. }
% \begin{tabular}{c|l|cccc|cccc}
\begin{tabularx}{\linewidth}{X<{\centering}|l|X<{\centering}X<{\centering}X<{\centering}X<{\centering}|X<{\centering}X<{\centering}|X<{\centering}X<{\centering}} \toprule
                                                                            & Method                        & RMSE$\downarrow$ & SSIM$\uparrow$  & PSNR$\uparrow$   & SD$\uparrow$     & Acc.$\uparrow$  & mIoU$\uparrow$  & $P_\text{alex}$ $\downarrow$ & $P_\text{squeeze}$ $\downarrow$ \\ \midrule
\multirow{3}{*}{\begin{tabular}[c]{@{}c@{}}CVACT \\ (Aligned)\end{tabular}} & Pix2Pix~\cite{isola2017image} & 49.15          & 0.3733          & 14.47          & 16.06          & 0.8318          & 0.2137          & 0.4506                       & 0.2921                          \\
                                                                            & XFork~\cite{regmi2018cross}   & 50.81          & 0.3701          & 14.17          & 15.89          & 0.8227          & 0.2123          & 0.4408                       & 0.2932                          \\
                                                                            & Ours                          & \textbf{48.23} & \textbf{0.4212} & \textbf{14.65} & \textbf{16.33} & \textbf{0.8353} & \textbf{0.2145} & \textbf{0.4099}              & \textbf{0.2701}                 \\ \midrule
\multirow{3}{*}{CVUSA}                                                      & Pix2Pix~\cite{isola2017image} & 56.11          & 0.2952          & 13.35          & 15.90          & 0.7742          & 0.2042          & 0.5037                       & 0.3774                          \\
                                                                            & XFork~\cite{regmi2018cross}   & 56.77          & 0.2926          & 13.25          & 15.80          & 0.7722          & 0.2039          & 0.5144                       & 0.3836                          \\
                                                                            & Ours                          & \textbf{53.67} & \textbf{0.3408} & \textbf{13.77} & \textbf{16.27} & \textbf{0.7831} & \textbf{0.2060} & \textbf{0.4824}              & \textbf{0.3577}               \\ \bottomrule  
\end{tabularx}
\label{tab: quantitative_stoa}
\end{table*}

\subsection{Implementation details}
In our implementation, we resize the input satellite image to $256\times 256$ pixels and set the output size of a street-view panorama to $128 \times 512$ pixels. 
We approximate the height of the street-view camera as $2$ meters with respect to the ground plane. 
The maximum height modeled by our multiplane image representation is $8$ meters.
We set the number of satellite-view MPI planes $N$ to $64$ in our experiments, with a half meter interval between planes. 
{\color{black}
The number of street-view MPI planes $M$ is also set to $64$.
The network is trained in an end-to-end manner with a batch size of 4.
We follow Pix2Pix's~\cite{isola2017image} use of the Adam optimizer~\cite{kingma2015adam} with a learning rate of $0.0002$ for both the generator and discriminator, and $\beta_1 = 0.5$, $\beta_2 =0.9999$. 
The source code is available at \url{https://github.com/shiyujiao/Sat2StrPanoramaSynthesis.git}. 

The street-view image rendering complexity is $\small {O}(H^{\text{str}}W^{\text{str}}M)$. The flops in $ \alpha^{\text{str}} \cdot C^{\text{str}}$ are $3H^{\text{str}}W^{\text{str}}M$, and the flops in $ (1-\alpha^{\text{str}}) \cdot I_\text{comp}$ are $4H^{\text{str}}W^{\text{str}}M$. 
Using an RTX 2080 TI, the maximum batch size is 12 for training and 64 for testing. The training time is around 12 hours for the CVACT dataset and 16 hours for the CVUSA dataset. 
On average, the computation time is 0.2s per synthesized image.
} 
\subsection{Evaluation metrics}

In this paper, we adopt various evaluation metrics for quantitative assessment. 
The low-level similarity measures includes {\color{black}root-mean-square error} (RMSE), {\color{black}structure similarity index measure} (SSIM), {\color{black}peak signal-to-noise ratio} (PSNR), and sharpness difference (SD). 
They evaluate the pixel-wise similarity between two images. 
{\color{black}However, in the cross-view synthesis task, it is hard to synthesize exactly the same target view image as the ground truth, due to the minimal overlap and seasonal change between input and target view images.
The colors between the generated and ground truth street-view images may be different, but they depict the same location. }
Thus, we further adopt the high-level perceptual similarity~\cite{zhang2018unreasonable} for the performance evaluation. 
The perceptual similarity evaluates the feature similarity of generated and real images. 
We employ the pretrained AlexNet~\cite{krizhevsky2012imagenet} and Squeeze~\cite{iandola2016squeezenet} networks as backbones for the evaluation, denoted as $P_\text{alex}$ and $P_\text{squeeze}$, respectively.

Additionally, we employ a pre-trained semantic classifier to measure the semantic difference between the real and generated images. The semantic classifier is trained on the CityScapes dataset~\cite{cordts2016cityscapes} and fine-tuned on the CVUSA dataset by using the lightweight RefineNet~\cite{nekrasov2018light} network. We report the pixel-wise accuracy (Acc.) and {\color{black} mean intersection over union} (mIoU), as in the work~\cite{long2015fully}.

\subsection{Comparison with existing methods}
We compare our method with Pix2Pix~\cite{isola2017image} and XFork~\cite{regmi2018cross}.
Pix2Pix is a well-known GAN-based network for image-to-image translation and has been widely used as the baseline for cross-view image synthesis~\cite{regmi2018cross,tang2019multi}.
XFork, proposed by Regmi and Borji~\cite{Regmi_2019_ICCV}, used semantic information as additional network guidance during training, generating the target image and semantic map simultaneously with a weight-shared decoder.
The authors also proposed another network, XSeq, which stacked two generators together to generate the target image and semantic map sequentially. We compare with XFork, which was shown to outperform XSeq~\cite{regmi2018cross}.
Selection GAN~\cite{tang2019multi} is another recent work on cross-view image synthesis. However, it assumed that a semantic segmentation of the street-view panorama was available during testing, which is different to our problem setting.
Our goal in this paper is to synthesize a street-view panorama from a satellite image without any information from the target domain.
Lu \etal~\cite{lu2020geometry} also exploited
geometric correspondences for satellite to street-view image synthesis. However, their work needs explicit height and semantic supervision for satellite images, which is not available for current 
accessible datasets (CVUSA and CVACT). 
Therefore, we cannot meaningfully compare with these two methods.

Table~\ref{tab: quantitative_stoa} presents the quantitative comparison results. As indicated in the table, our method achieves consistently better results on all quantitative evaluation metrics. 
Figure~\ref{fig:Qualitative comparison} provides some qualitative visualizations from the CVACT (aligned) and CVUSA datasets.
As indicated by the results, our pipeline generally produces more natural images, which can be observed from the first two examples in Figure~\ref{fig:visualization on CVACT} and the first example in Figure~\ref{fig:visualization on CVUSA}, where the images generated by Pix2Pix and XFork confuse some regions and inpaint them with artifacts. 
In particular, our method generates street-view panoramas that are more geometrically-consistent with respect to the input satellite images.

\begin{table*}[!ht!]
\setlength{\abovecaptionskip}{0pt}
\setlength{\belowcaptionskip}{0pt}
\setlength{\tabcolsep}{10pt}
\centering
\caption{\small Performance comparison on the aligned and unaligned CVACT dataset (both training and testing). Here, the {\ul \textit{underlined}} number indicates the performance change (in percent) between 
a model trained on unaligned image pairs the same model trained on aligned image pairs. 
The arrow ``$\downarrow$'' indicates performance decrease and ``$\uparrow$'' indicates the performance increase. }
\begin{tabularx}{\linewidth}{l|l|X<{\centering}X<{\centering}X<{\centering}X<{\centering}|X<{\centering}X<{\centering}X<{\centering}X<{\centering}}
\toprule
% \begin{tabular}{cccccccccc}
\multirow{2}{*}{}        &  \multicolumn{1}{c|}{\multirow{2}{*}{\diagbox{Train}{Test}}} & \multicolumn{4}{c|}{Aligned}                                                                                                                   & \multicolumn{4}{c}{Unaligned}                                                                                                                 \\
                         &                   & RMSE$\downarrow$                  & SSIM$\uparrow$                    & PSNR$\uparrow$                    & SD$\uparrow$                      & RMSE$\downarrow$                  & SSIM$\uparrow$                    & PSNR$\uparrow$                    & SD$\uparrow$                      \\ \midrule
\multirow{3}{*}{Pix2Pix~\cite{isola2017image}} & Unaligned         & 50.21                           & 0.3704                            & 14.29                           & 16.03                           & 51.79                           & 0.3563                            & 14.03                           & 15.88                           \\
                         & Aligned           & 49.15                           & 0.3733                            & 14.47                           & 16.06                           & 51.39                           & 0.3544                            & 14.10                           & 15.91                           \\
                         & Change (\%)        & {\ul \textit{2.122$\downarrow$}} & {\ul \textit{0.7704$\downarrow$}} & {\ul \textit{1.321$\downarrow$}} & {\ul \textit{0.1874$\downarrow$}} & {\ul \textit{0.7730$\downarrow$}} & {\ul \textit{0.5100$\uparrow$}}   & {\ul \textit{0.4762$\downarrow$}} & {\ul \textit{0.2044$\downarrow$}} \\ \midrule
\multirow{3}{*}{XFork~\cite{regmi2018cross}}   & Unaligned         & 51.36                           & 0.3638                            & 14.10                           & 15.80                           & 53.06                           & 0.3485                            & 13.83                           & 15.65                           \\
                         & Aligned           & 50.81                           & 0.3701                            & 14.17                           & 15.89                           & 53.01                           & 0.3497                            & 13.82                           & 15.71                           \\
                         & Change (\%)        & {\ul \textit{1.068$\downarrow$}} & {\ul \textit{1.744$\downarrow$}} & {\ul \textit{0.5450$\downarrow$}} & {\ul \textit{0.5903$\downarrow$}} & {\ul \textit{0.0830$\downarrow$}} & {\ul \textit{0.3340$\downarrow$}} & {\ul \textit{0.0736$\uparrow$}}   & {\ul \textit{0.3745$\downarrow$}} \\ \midrule
\multirow{3}{*}{Ours}    & Unaligned         & 48.37                           & 0.4210                            & 14.62                           & 16.27                           & 49.82                           & 0.4014                            & 14.38                           & 16.24                           \\
                         & Aligned           & 48.23                           & 0.4212                            & 14.65                           & 16.33                           & 50.66                           & 0.3901                            & 14.24                           & 16.07                           \\
                         & Change (\%)        & {\ul \textit{0.2910$\downarrow$}} & {\ul \textit{0.0411$\downarrow$}} & {\ul \textit{0.2481$\downarrow$}} & {\ul \textit{0.3379$\downarrow$}} & {\ul \textit{1.701$\uparrow$}}   & {\ul \textit{2.813$\uparrow$}}   & {\ul \textit{0.9244$\uparrow$}}   & {\ul \textit{1.011$\uparrow$}}  \\ \bottomrule
\end{tabularx}
\label{tab: aligned_unaligned}
\end{table*}

As shown in the first example of Figure~\ref{fig:visualization on CVACT}, there is a roundabout in the input satellite image, which can also be observed from the corresponding street-view panorama (Ground Truth). 
Our method successfully projects the estimated scene geometry and recover this structure.
The fully black-box networks Pix2Pix and XFork fail to learn the geometric transformations and generate unnatural synthesized images.
The same phenomenon can be observed in the second example in Figure~\ref{fig:visualization on CVACT}.
Notably, our method can recover the lane line structure from the satellite image, as shown in the third and fourth example in Figure~\ref{fig:visualization on CVACT}.
Such lane line structure is challenging for a fully black-box model to generate, since different locations will have different structures and lane lines occupy a relatively small region in both input and output images.

Figure~\ref{fig:visualization on CVUSA} presents some challenging cases with a variety of complex scene structures from the CVUSA dataset.
It can be seen that our method recovers the geometric structures from satellite images, while generating diverse and appropriate visual features.
In contrast, the generated images from Pix2Pix and XFork are much more uniform for the first two examples, with two main roads uniformly distributed and other regions in-painted with grass.
For a more complex case, the third example in Figure~\ref{fig:visualization on CVUSA}, Pix2Pix and XFork fail entirely while our method is more successful at recovering the road structure from the input satellite image.
Preserving the correct road topology is very important in many applications, such as autonomous driving.
The last example in Figure~\ref{fig:visualization on CVUSA} presents the most challenging case.
The geometric structure of the input satellite image is complicated, and this kind of image (urban jungle) is rare in the training set. Pix2Pix and XFork hallucinate the ground-level scene, while our approach generates street-view panoramas that are more geometrically consistent with the ground truth.
More specifically,
our method 
restores (roughly) the ground and building geometry from the satellite image.

\begin{table*}[]
\setlength{\abovecaptionskip}{0pt}
\setlength{\belowcaptionskip}{0pt}
\centering
\caption{\small Ablation study on the CVACT (Aligned) and the CVUSA dataset. 
}
\begin{tabular}{l|l|c|cccc|cccc}
\toprule
\multirow{2}{*}{}            & \multirow{2}{*}{Method} & \multirow{2}{*}{Params} & \multicolumn{4}{c}{CVACT (Aligned)}                                      & \multicolumn{4}{c}{CVUSA}                                                \\ 
                             &                       &                         & RMSE$\downarrow$     & SSIM $\uparrow$ & PSNR $\uparrow$  & SD $\uparrow$  & RMSE $\downarrow$  & SSIM $\uparrow$   & PSNR $\uparrow$    & SD $\uparrow$               \\ \midrule
\multirow{2}{*}{Handcrafted} & Polar                                       & -                       & 82.08          & 0.1503          & 9.941           & 13.83          & 74.50          & 0.1427          & 10.79          & 14.46          \\
                             & Projection                                  & -                       & 118.0         & 0.1955          & 6.725           & 17.18          & 118.2         & 0.0742          & 6.737           & 16.22          \\ \midrule
\multirow{7}{*}{Deep}        & Ours w/o S2SP                               & 33.46M                  & 48.63          & 0.4050          & 14.57          & 16.15          & 54.05          & 0.3197          & 13.70          & 16.08          \\
                             & Ours w/o height (polar)                     & 33.46M                  & 49.46          & 0.3962          & 14.42          & 16.13          & 54.78          & 0.3122          & 13.57          & 16.00          \\
                             & Ours w/o height (projection)                & 33.46M                  & 49.08          & 0.4068          & 14.48          & 16.18          & 54.49          & 0.3301          & 13.63          & 16.11          \\
                             & Ours w/o projection                         & 33.62M                  & 49.47          & 0.4074          & 14.43          & 16.14          & 53.96          & 0.3216          & 13.72          & 16.12          \\
                             & Ours w/o $\mathcal L_1$                     & 33.62M                  & 48.76          & 0.4174          & 14.55          & 16.20          & 59.25          & 0.1512          & 12.79          & 10.56          \\
                             & Ours w/o $\mathcal L_{\text{per}}$          & 33.62M                  & 50.96          & 0.3973          & 14.18          & 16.10          & 55.43          & 0.3230          & 13.47          & 16.10          \\
                             & Ours                                        & 33.62M                  & \textbf{48.23} & \textbf{0.4212} & \textbf{14.65} & \textbf{16.33} & \textbf{53.67} & \textbf{0.3408} & \textbf{13.77} & \textbf{16.27} \\ \bottomrule
\end{tabular}
\label{ablation}
\end{table*}

{\color{black}
\subsection{Comparison on aligned or unaligned datasets}
In this section we evaluate the performance of algorithms on the aligned and unaligned CVACT test sets with models trained on the aligned and unaligned training sets. 
We first consider the results on the aligned test set, which are more reliable measures of image synthesis performance.
As shown in Table~2, all methods perform worse when trained on unaligned image pairs compared to aligned image pairs.
This is expected since the supervision signal is noisier during training.
Our method is the least sensitive, likely due to the explicit assumption in the projection module that the viewpoint of the synthesized image is strictly at the center of the satellite image.

We also evaluate the models on the unaligned test set. 
All models perform worse on this test set compared to the aligned test set, regardless of whether training data is aligned or not, since the ``ground truth'' is not correct or predictable.
The performance of the Pix2Pix and XFork models decreases when trained on unaligned data, which is consistent with the results on the aligned test set. 
However, the magnitude of the decrease is much smaller.
Since these models do not have an explicit alignment rule, the alignment of the training data does not affect the evaluations on unaligned data too much.
In contrast, 
the performance of our method is inferior when trained on aligned data compared to unaligned data, 
because the explicit alignment rule in our method reduces the network's capacity to learn any misalignment bias.
}

\subsection{Ablation study}

\begin{figure*}[!ht]
    \setlength{\abovecaptionskip}{0pt}
    \setlength{\belowcaptionskip}{0pt}
    \centering
    \subfloat[ \small Satellite \protect\\\centering{Image}]{
    \centering
    \begin{minipage}{0.1\linewidth}
    \includegraphics[width=\linewidth]{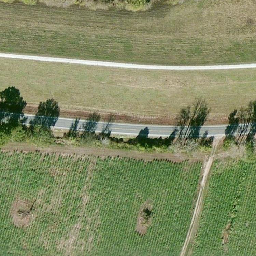}
    \includegraphics[width=\linewidth]{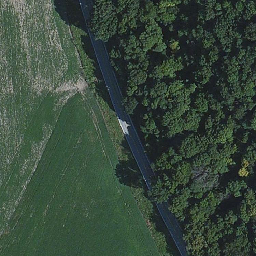}
    \includegraphics[width=\linewidth]{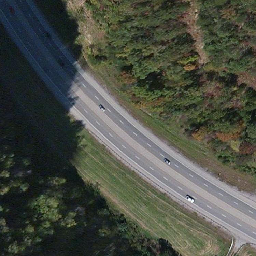}
    \end{minipage}
    }
    \subfloat[ \small {\color{black}Height} \protect\\\centering{{\color{black}Map}}]{
    \centering
    \begin{minipage}{0.1\linewidth}
    \includegraphics[width=\linewidth]{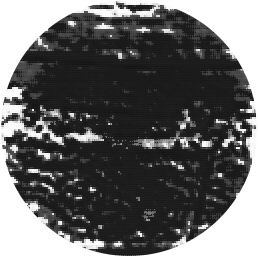}
    \includegraphics[width=\linewidth]{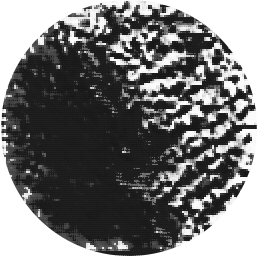}
    \includegraphics[width=\linewidth]{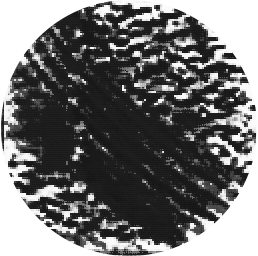}
    \end{minipage}
    \label{ablation_height}
    }
    \subfloat[ \small Projected Satellite Image \protect\\\centering{(Intermediate)}]{
    \centering
    \begin{minipage}{0.245\linewidth}
    \includegraphics[width=\linewidth, height=0.4\linewidth]{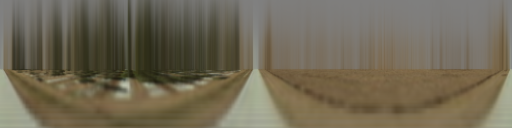}
    \includegraphics[width=\linewidth, height=0.4\linewidth]{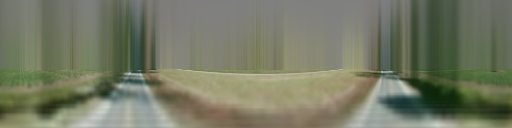}
    \includegraphics[width=\linewidth, height=0.4\linewidth]{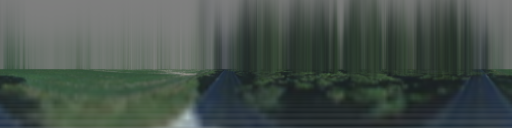}
    \end{minipage}
    \label{projected satellite (additional)}
    }
    \subfloat[ \small Generated Street-View  \protect\\\centering{Panorama (Final)}]{
    \centering
    \begin{minipage}{0.245\linewidth}
    \includegraphics[width=\linewidth, height=0.4\linewidth]{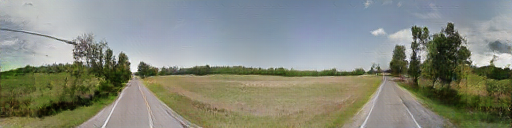}
    \includegraphics[width=\linewidth, height=0.4\linewidth]{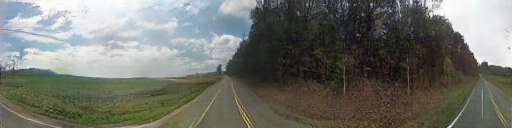}
    \includegraphics[width=\linewidth, height=0.4\linewidth]{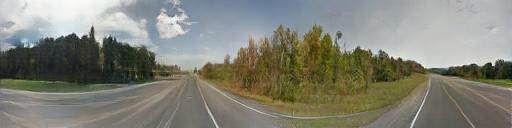}
    \end{minipage}
    }
    \subfloat[ \small Real Street-View Panorama \protect\\\centering{(Ground Truth)}]{
    \centering
    \begin{minipage}{0.245\linewidth}
    \includegraphics[width=\linewidth, height=0.4\linewidth]{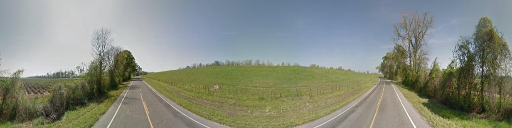}
    \includegraphics[width=\linewidth, height=0.4\linewidth]{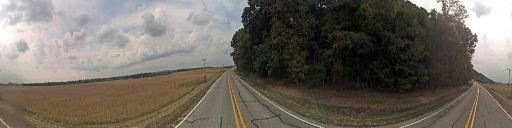}
    \includegraphics[width=\linewidth, height=0.4\linewidth]{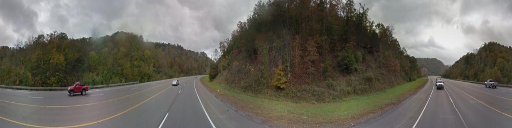}
    \end{minipage}
    }
    \caption{\small  Additional qualitative results with {\color{black} height maps (lighter is higher)}, projected satellite images generated by our S2SP module, and synthesized images generated by our entire pipeline.
    }
    \label{fig:abalation_visualization}
\end{figure*}

\begin{figure*}[!ht]
\setlength{\abovecaptionskip}{0pt}
    \setlength{\belowcaptionskip}{0pt}
\centering
% \subfloat[CVACT]{
\subfloat[\small Satellite Image]{
\centering
    \begin{minipage}{0.13\linewidth}
    \centering
    % \begin{minipage}[t][\linewidth][t]{\linewidth}
    \includegraphics[width=0.83077\linewidth]{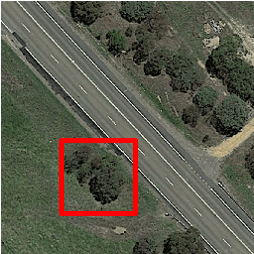}
    % \end{minipage}
    % \begin{minipage}[t][\linewidth][t]{\linewidth}
    \includegraphics[width=0.83077\linewidth]{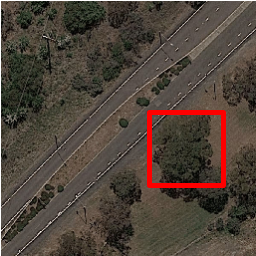}
    % \end{minipage}
    % \centerline{\small Satellite Image}\\
    \end{minipage}
    }\hfill
    \subfloat[\small Ours w/o Height (Projection)]{
    \begin{minipage}{0.27\linewidth}
    % \begin{minipage}[t][\linewidth][t]{\linewidth}
    \includegraphics[width=\linewidth, height=0.4\linewidth]{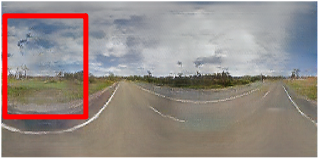}
    % \end{minipage}
    % \begin{minipage}[t][\linewidth][t]{\linewidth}
    \includegraphics[width=\linewidth, height=0.4\linewidth]{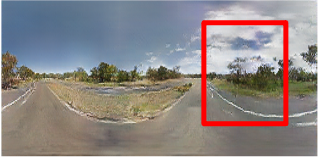}
    % \end{minipage}
    % \centerline{\small Ours w/o height (projection)}\\
    % \centerline{\small (projection) w_ $L_\text{per}$}
    \end{minipage}
    }\hfill
    \subfloat[\small Ours]{
    \begin{minipage}{0.27\linewidth}
    % \begin{minipage}[t][\linewidth][t]{\linewidth}
    \includegraphics[width=\linewidth, height=0.4\linewidth]{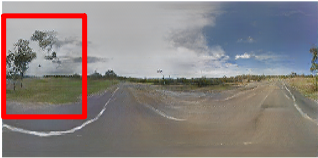}
    % \end{minipage}
    % \begin{minipage}[t][\linewidth][t]{\linewidth}
    \includegraphics[width=\linewidth, height=0.4\linewidth]{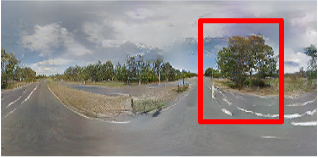}
    % \centerline{\small Ours }\\
    % \centerline{\small $L_\text{per}$}
    \end{minipage}
    }\hfill
    \subfloat[\small Ground Truth]{
    \begin{minipage}{0.27\linewidth}
    % \begin{minipage}[t][\linewidth][t]{\linewidth}
    \includegraphics[width=\linewidth, height=0.4\linewidth]{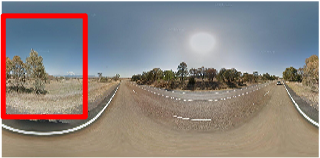}
    % \end{minipage}
    % \begin{minipage}[t][\linewidth][t]{\linewidth}
    \includegraphics[width=\linewidth, height=0.4\linewidth]{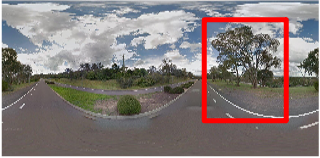}
    % \centerline{\small Ground Truth}\\
    \end{minipage}
    }
    % \label{fig:visualization on CVACT}
    % }\\
    \caption{\small Qualitative comparison of images synthesized by our method with learned height maps (`Ours') and with height maps fixed to zero (`Ours w/o Height (Projection)').
    % The first two samples are from the CVACT dataset and the last two samples are from the CVUSA dataset.
    }
    \label{fig:Qualitative abla1}
\end{figure*}

\begin{figure*}[!htbp]
    \setlength{\abovecaptionskip}{0pt}
    \setlength{\belowcaptionskip}{0pt}
    \centering
    \subfloat[ \small Satellite Image]{
    \centering
    \begin{minipage}{0.13\linewidth}
    \centering
    \includegraphics[width=0.83077\linewidth]{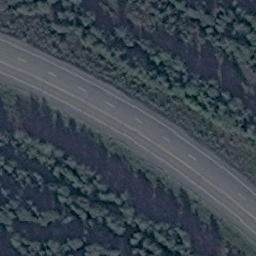}
    \includegraphics[width=0.83077\linewidth]{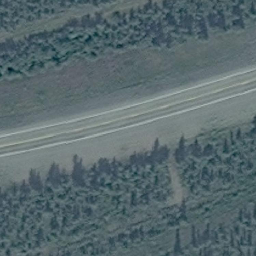}
    \end{minipage}
    } \hfill
    \subfloat[ \small Ours w/o Height (Projection) ]{
    \centering
    \begin{minipage}{0.27\linewidth}
    \includegraphics[width=\linewidth, height=0.4\linewidth]{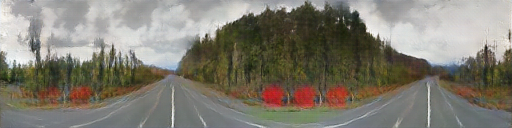}
    \includegraphics[width=\linewidth, height=0.4\linewidth]{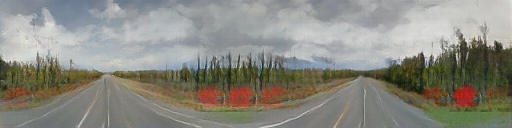}
    \end{minipage}
    }\hfill
    \subfloat[ \small Ours ]{
    \centering
    \begin{minipage}{0.27\linewidth}
    \includegraphics[width=\linewidth, height=0.4\linewidth]{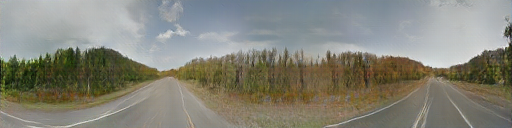}
    \includegraphics[width=\linewidth, height=0.4\linewidth]{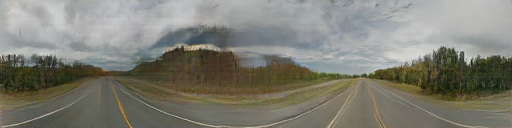}
    \end{minipage}
    }\hfill
    \subfloat[ \small Ground Truth]{
    \centering
    \begin{minipage}{0.27\linewidth}
    \includegraphics[width=\linewidth, height=0.4\linewidth]{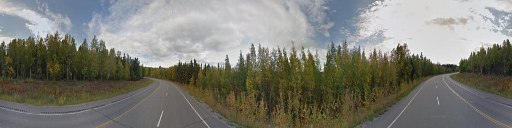}
    \includegraphics[width=\linewidth, height=0.4\linewidth]{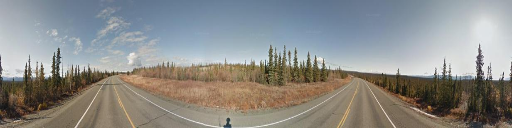}
    \end{minipage}
    }
    \caption{\small Qualitative comparison of images synthesized from low-quality satellite images.}
    \label{fig:Qualitative abla2}
\end{figure*}

In this section, experiments are carried out to validate the importance of each component in our framework.
We divide our experiments into two groups: handcrafted methods and deep learning methods.
For the handcrafted features, we compare with the polar transform~\cite{shi2019spatial} and our satellite to street-view image projection without height estimation, denoted as ``Polar'' and ``Projection'', respectively.
Both of these methods assume that pixels in the satellite image lie on the ground plane. 
``Polar'' is a simple approximation for cross-view image alignment, while ``Projection'' establishes the real geometric correspondences for pixels which have ground-level height in a satellite and street-view image pair.

For deep learning approaches, all of the comparison algorithms employ the Pix2Pix network as the generator backbone. 
We first remove the whole S2SP module from our pipeline, denoted as ``Ours w/o S2SP''.
In this pipeline, only the generator backbone is retained and the generator is conditioned on the original satellite image.
Next, we remove the height estimation block from our pipeline and assume that all the pixels have the same height as the ground plane (zero height). 
Two types of generator conditionings are investigated with this baseline: the polar-transformed images and the perspective-projected images, denoted as ``Ours w/o height (polar)'' and ``Ours w/o height (projection)'', respectively.
We also investigate the usefulness of the geometry projection block by replacing it with a simpler approach: a soft argmax operation to convert the estimated height probability distribution into a single height map, which is concatenated with the satellite image to condition the generator, denoted as ``Ours w/o projection''.
{
\color{black}
Finally, we study the influence of the $\mathcal L_1$ and $\mathcal L_{\text{per}}$ losses by removing them from the total image reconstruction loss, denoted as ``Ours w/o $\mathcal L_1$'' and ``Ours w/o $\mathcal L_{\text{per}}$'', respectively. 
}

The ablation study is presented in Table~\ref{ablation}.
The handcrafted methods perform significantly worse than the deep learning methods, indicating that the satellite to street-view transformation is too complex to be modeled by simple approximations.
For the deep learning methods, our whole pipeline consistently achieves the best results on all evaluation metrics, indicating that every component is important for the success of the model.

{\color{black}
Regarding the $\mathcal L_1$ and $\mathcal L_{\text{per}}$ losses, we found that using $\mathcal L_{\text{per}}$ significantly improves the quality of the synthesized images, as indicated by the last two rows of Table~\ref{ablation}. 
This loss provides higher-level feature similarity guidance to the network in addition to the pixel-wise $\mathcal L_1$ color difference during training. 
This is especially important in the cross-view synthesis task, since the ground truth target-view images in the training set are not captured at the same time as the input-view images. 
Purely using the $\mathcal L_1$ color loss makes it difficult for the network to learn the significant cross-view transformations. 
However, discarding the $\mathcal L_1$ loss also impairs the performance. 
This can be observed from the third last row of Table~\ref{ablation}, where the performance of ``Ours w/o $\mathcal L_1$'' drops significantly on the CVUSA dataset. 
The synthesized images of ``Ours w/o $\mathcal L_1$'' on the CVUSA dataset are very blurry, suggesting that the $\mathcal L_1$ loss is responsible for producing sharper images.
}

\begin{figure}[ht!]
\setlength{\abovecaptionskip}{0pt}
\setlength{\belowcaptionskip}{0pt}
    \centering
    \begin{minipage}{0.24\linewidth}
        \includegraphics[width=\linewidth]{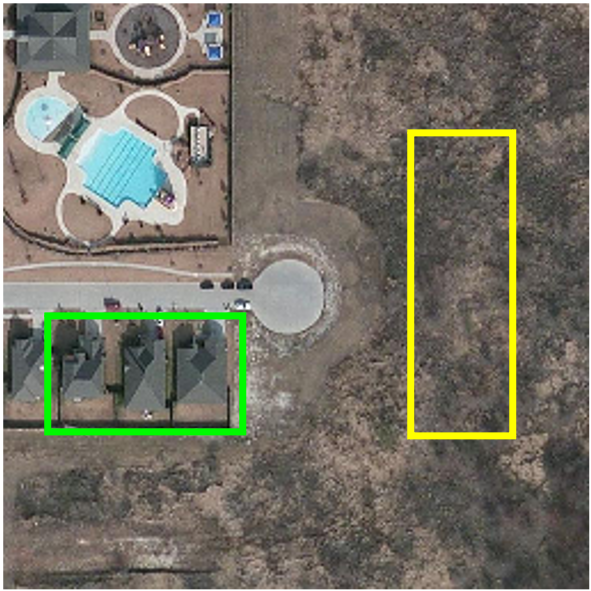}
        \includegraphics[width=\linewidth]{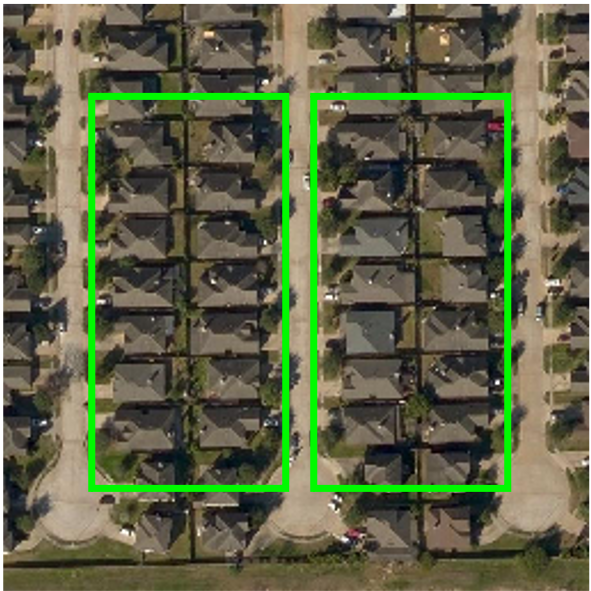}
        \includegraphics[width=\linewidth]{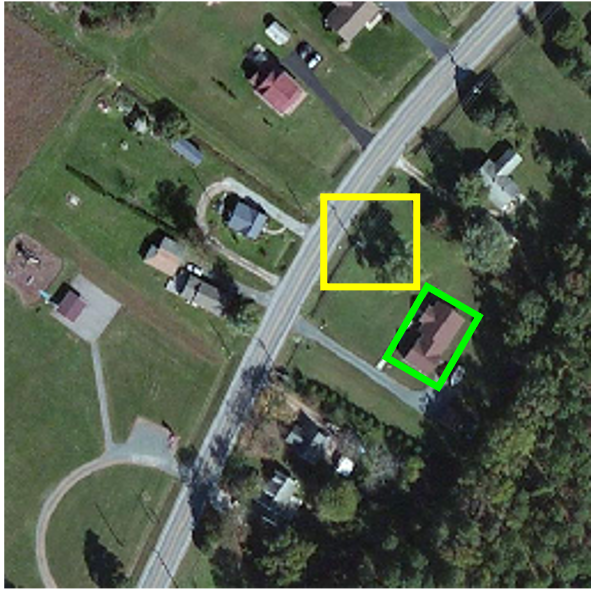}
        \centerline{\small Satellite Image}
    \end{minipage}
    \begin{minipage}{0.24\linewidth}
        \includegraphics[width=\linewidth]{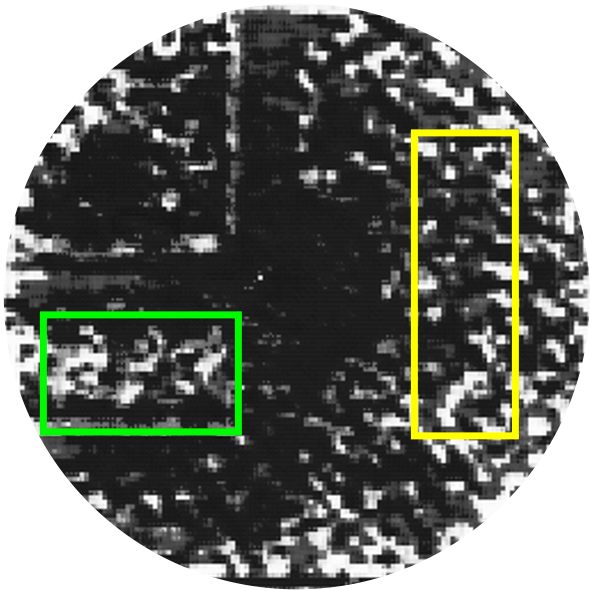}
        \includegraphics[width=\linewidth]{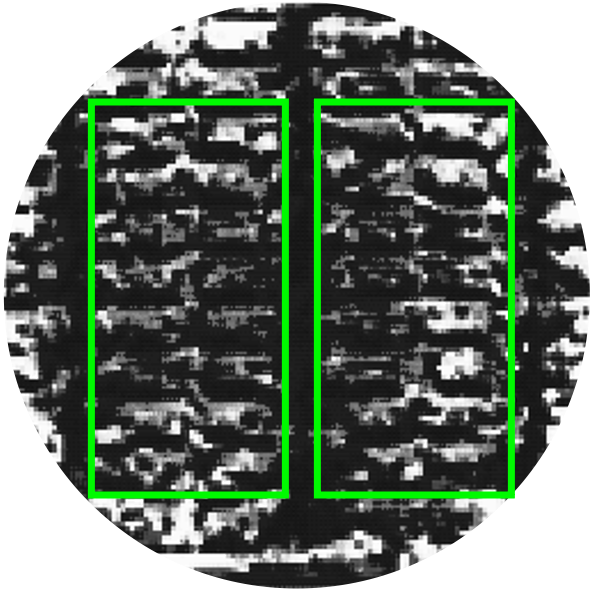}
        \includegraphics[width=\linewidth]{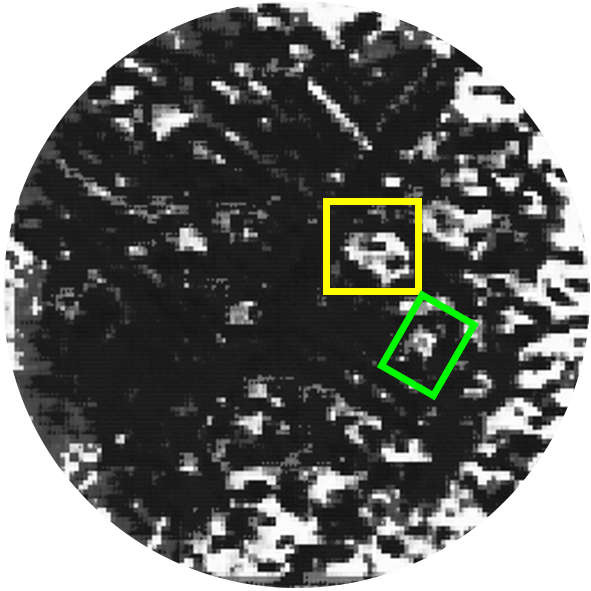}
        \centerline{\small Height Map}
    \end{minipage}
    \hfill
    \begin{minipage}{0.24\linewidth}
        \includegraphics[width=\linewidth]{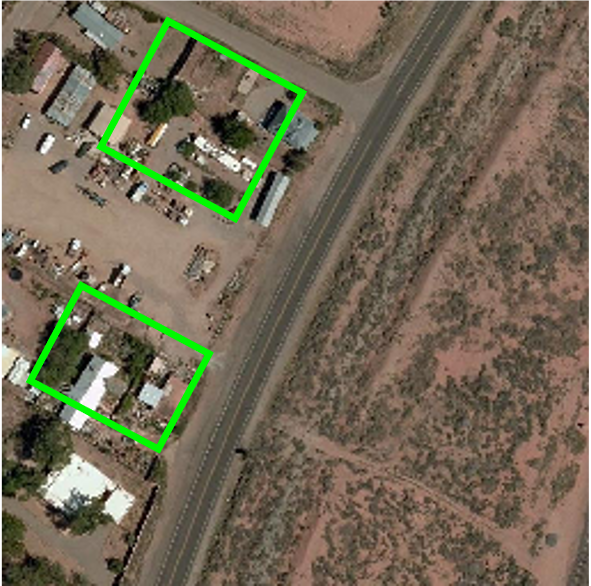}
        \includegraphics[width=\linewidth]{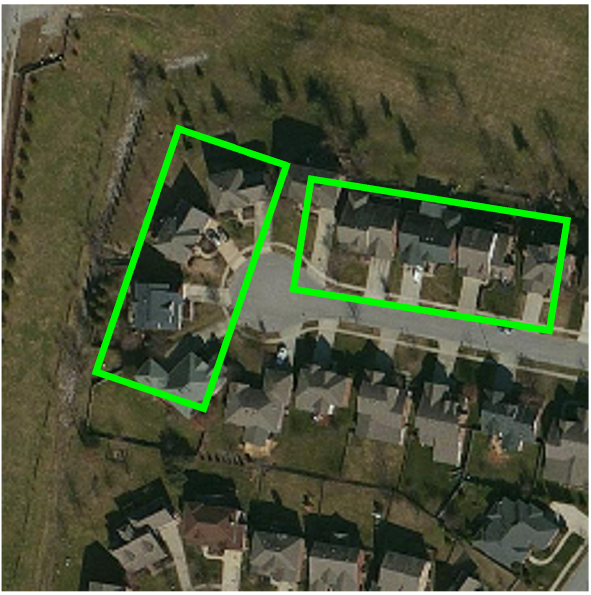}
        \includegraphics[width=\linewidth]{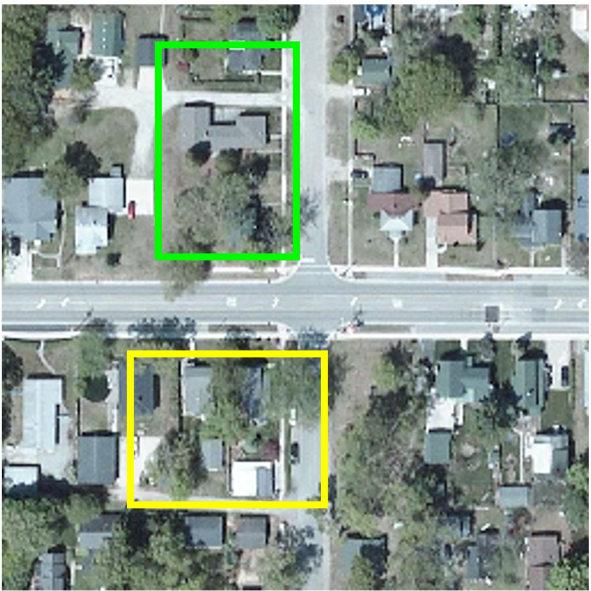}
        \centerline{\small Satellite Image}
    \end{minipage}
    \begin{minipage}{0.24\linewidth}
        \includegraphics[width=\linewidth]{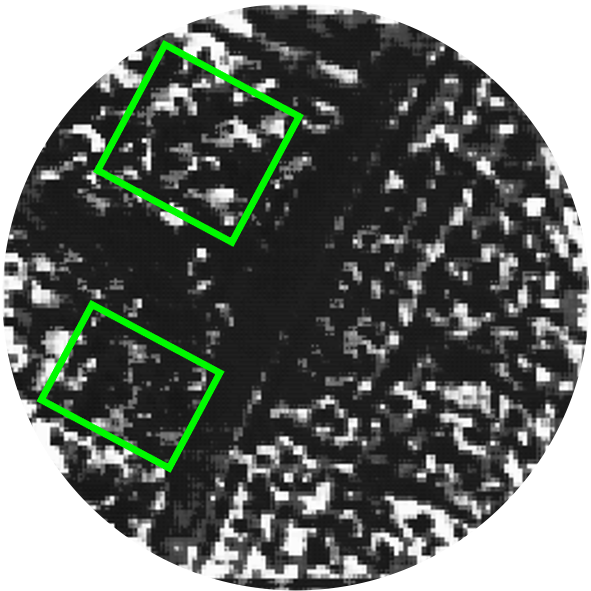}
        \includegraphics[width=\linewidth]{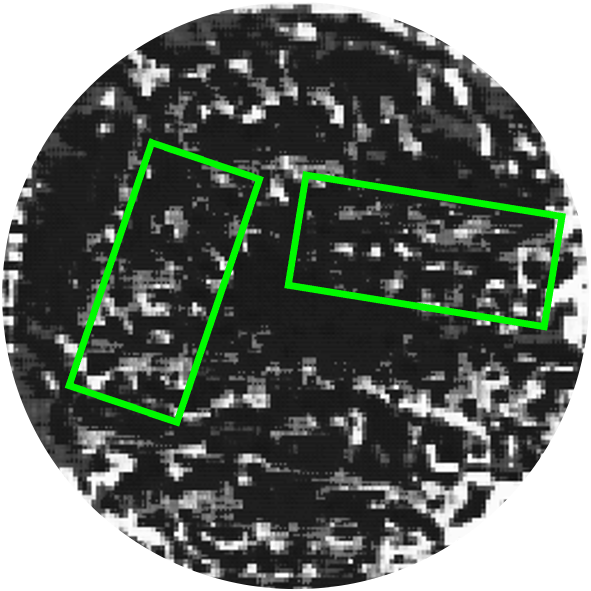}
        \includegraphics[width=\linewidth]{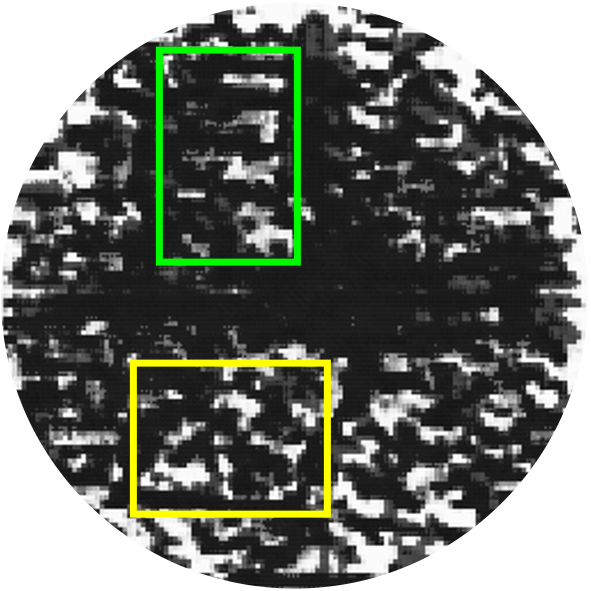}
        \centerline{\small Height Map}
    \end{minipage}
    \caption{Additional estimated height maps for complex scenes. 
    Regions with greater heights (\ie, buildings and trees) are annotated with rectangles (buildings as green and trees as yellow).
    }
    \label{fig:additional_heights}
\end{figure}

\begin{figure*}[!htbp]
\setlength{\abovecaptionskip}{0pt}
\setlength{\belowcaptionskip}{0pt}
    \centering
    \subfloat[ \small Satellite-view MPI]{
    \parbox[][4.5cm][c]{0.25\linewidth}{
    \centering
    \includegraphics[width=\linewidth]{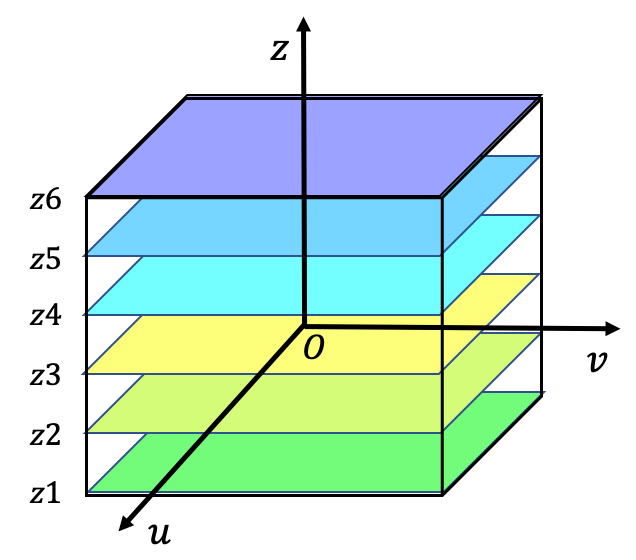}
    \label{aer_MPI_height}
    }
    }
    \hspace{2em}
    \subfloat[ \small Viewing rays]{
    \parbox[][4.5cm][c]{0.25\linewidth}{
    \centering
    \includegraphics[width=\linewidth]{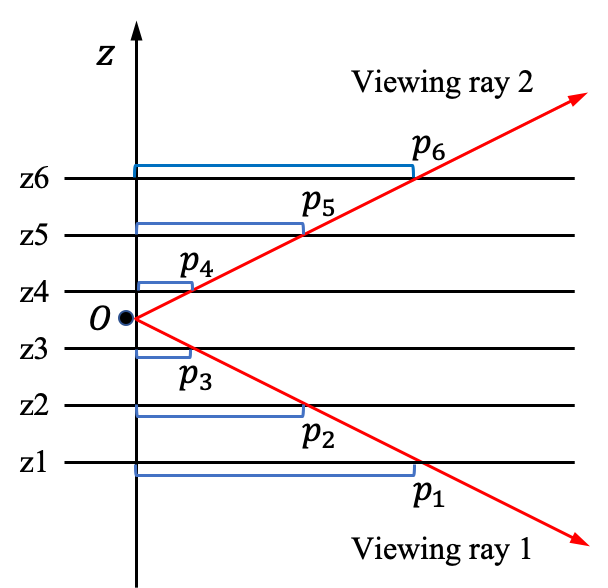}
    \label{viewing_ray}
    }
    }
    \hspace{2em}
    \subfloat[ \small Plane reorder]{
    \parbox[][4.5cm][c]{0.25\linewidth}{
    \centering
    \includegraphics[width=\linewidth]{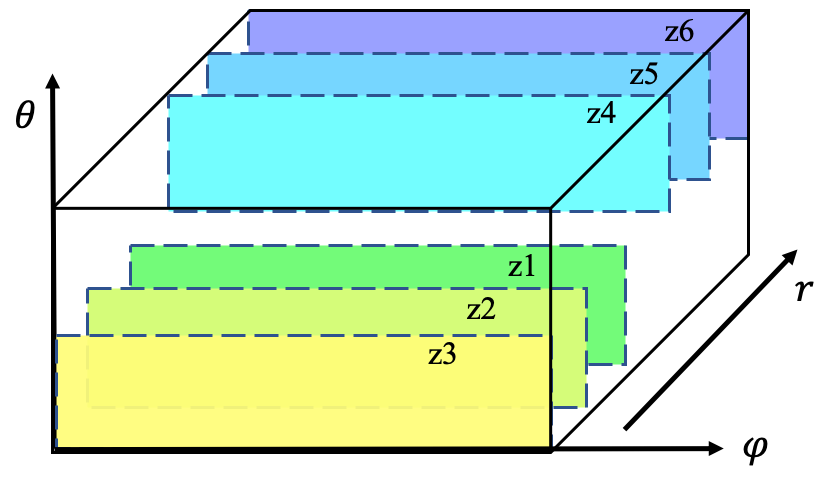}
    \label{plane_reorder}
    }
    }
    \caption{\small ``Height-wise'' satellite to street-view image projection method. We directly project each of the image planes in an satellite-view MPI (a) to the street viewpoint, and then adjust the order of projected planes according to the principle illustrated in (b). The result is a volumetric scene representation that sorts points from far to near along viewing rays (c).}
    \label{fig:height-wise projection}
\end{figure*}

\begin{table*}[]
\setlength{\abovecaptionskip}{0pt}
\setlength{\belowcaptionskip}{0pt}
\setlength{\tabcolsep}{10pt}
\centering
\caption{\small Performance comparison as the number of planes $N$ varies on the CVACT (aligned) and the CVUSA dataset.}
\begin{tabularx}{\linewidth}{X<{\centering}|X<{\centering}|X<{\centering}X<{\centering}X<{\centering}X<{\centering}|X<{\centering}X<{\centering}X<{\centering}X<{\centering}}
\toprule
% \begin{tabular}{clcccccccc}
\multirow{2}{*}{$N$} & \multirow{2}{*}{Params} & \multicolumn{4}{c|}{CVACT (Aligned)}                                      & \multicolumn{4}{c}{CVUSA}                                             \\
                     &                         & RMSE$\downarrow$ & SSIM$\uparrow$  & PSNR$\uparrow$   & SD$\uparrow$     & RMSE$\downarrow$ & SSIM$\uparrow$ & PSNR$\uparrow$  & SD$\uparrow$    \\ \midrule
1                    & 33.46M                  & 49.08          & 0.4068          & 14.48          & 16.18          & 54.49           & 0.3301          & 13.63          & 16.11          \\
16                   & 33.62M                  & \textbf{47.89} & 0.4178          & \textbf{14.71} & 16.29          & 53.73           & 0.3316          & 13.76          & 16.21          \\
32                   & 33.62M                  & 48.91          & 0.4140          & 14.52          & \textbf{16.40} & 54.58           & 0.3249          & 13.61          & 16.10          \\
64                   & 33.63M                  & 48.23          & \textbf{0.4212} & 14.65          & 16.33          & \textbf{53.67}  & \textbf{0.3408} & \textbf{13.77} & \textbf{16.27} \\ \bottomrule
\end{tabularx}
\label{ablation_N}
\end{table*}

\begin{table*}[]
\setlength{\abovecaptionskip}{0pt}
\setlength{\belowcaptionskip}{0pt}
\centering
\caption{\small Performance investigation with different satellite to street-view projection approaches on the CVACT (aligned) dataset.}
\begin{tabularx}{\linewidth}{c|X<{\centering}|X<{\centering}|X<{\centering}X<{\centering}X<{\centering}X<{\centering}|X<{\centering}X<{\centering}X<{\centering}X<{\centering}}
\toprule
\multirow{2}{*}{}            & \multirow{2}{*}{} & \multirow{2}{*}{Params} & \multicolumn{4}{c}{CVACT (Aligned)}                                      & \multicolumn{4}{c}{CVUSA}                                                \\
                             &                   &                         & RMSE$\downarrow$ & SSIM$\uparrow$  & PSNR$\uparrow$   & SD$\uparrow$     & RMSE$\downarrow$ & SSIM$\uparrow$  & PSNR$\uparrow$   & SD$\uparrow$     \\ \midrule
\multirow{2}{*}{height-wise} & volume            & 33.82M                  & 48.00          & \textbf{0.4231} & 14.68          & 16.35          & 53.80          & 0.3276          & 13.74          & 16.18          \\
                             & image             & 33.63M                  & \textbf{47.72} & 0.4134          & \textbf{14.73} & 16.17          & 54.15          & 0.3301          & 13.69          & 16.12          \\ \midrule
\multirow{2}{*}{depth-wise}  & volume            & 33.82M                  & 48.23          & 0.4222          & 14.64          & \textbf{16.40} & 53.93          & 0.3337          & 13.73          & 16.17          \\
                             & image             & 33.63M                  & 48.23          & 0.4212          & 14.65          & 16.33          & \textbf{53.67} & \textbf{0.3408} & \textbf{13.77} & \textbf{16.27} \\ \bottomrule
\end{tabularx}
\label{ablation_a2g}
\end{table*}

\smallskip
{\color{black}
\noindent \textbf{Height estimation. } 
The ablation labelled as ``Ours w/o height (projection)'' is a variant of our method with a single MPI plane ($N=1$) corresponding to the ground plane.}
Generally, ``Ours w/o height (projection)'' recovers the ground structure of a scene from a satellite image, while scene objects with heights higher than the ground plane are hallucinated. 
This phenomenon can be easily observed from Figure~\ref{fig:Qualitative abla1}.
As shown in the two examples, the trees annotated in the satellite images have similar colors to their surrounding region. 
``Ours w/o height (projection)'' does not recognize the trees and the corresponding regions in the synthesized images are in-painted with grass.
In contrast, our whole pipeline with height estimation successfully distinguishes objects with different heights and the generated images are more geometrically consistent with the input satellite images. 
Figure~\ref{fig:Qualitative abla2} presents some examples on a special case where the input satellite images are of low quality.
As ``Ours w/o height (projection)'' tries to hallucinate the scenes for objects with heights higher than the ground plane, its generalization ability on these low quality input images is poor and the generated images have many artifacts.
Instead, our whole pipeline with the guidance of height estimation has more geometric clues, and thus the synthesized images are more natural.

{\color{black}
In Figure~\ref{fig:abalation_visualization}, we provide some additional qualitative results of our method, including the estimated height maps and the projected images from the S2SP module. 
We crop the height maps to the inscribed circle, since information outside that region is never used and receives no supervision. 
Figure~\ref{fig:additional_heights} provides more examples of estimated height maps for complex scenes. We highlight some regions of interest with rectangles, especially areas with greater heights.
}

The estimated height maps are coarse and sometimes inaccurate, due to the lack of explicit height supervision (with objects behind occluders having no supervision at all) and the use of an approximated camera height. However, misestimated heights can be tolerated by our subsequent generator network. Generally, our height estimation block is generally able to learn the statistical height distribution of different types of objects, \eg, trees are relatively higher than roads. The camera heights in the CVUSA and CVACT are approximated as $2$ meters in our implementation.

\smallskip
\noindent\textbf{Influence of the number of height planes.}
We investigate the performance of our method with different numbers of planes $N$ in the satellite-view MPI. 
The results are presented in Table~\ref{ablation_N}.
While increasing the number of height planes above one increases performance, further increases do not provide significant improvements. 
This may be because there is no explicit height supervision for the height estimation block and so the network learns the statistical height distribution of particular objects.
Therefore, beyond a certain point, the density of sampling along the height dimension may not make a difference.
We expect that a larger value of $N$ would be more advantageous when height supervision is available.

\subsection{Other satellite to street-view image projection methods}

In our S2SP module, we convert the satellite-view MPI to a street-view MPI by unrolling and stretching the concentric cylinders. 
In this section, we propose and discuss another satellite to street-view image projection method: 
directly projecting each of the image planes in the satellite-view MPI to the street viewpoint (from Figure~\ref{aer_MPI_height} to Figure~\ref{plane_reorder}).

Let $(u, v, z)$ denote the source coordinates of points in the image plane of the satellite-view MPI ($z$ is constant for each plane), and $(\theta, \phi)$ denote the coordinates of panorama projection rays viewed at the ground level.
The projection between source and target coordinates for each plane in Figure~\ref{aer_MPI_height} can be expressed by Eq.~\ref{sat2str}.
As indicated by the equation, for fixed $\theta$, the distance between the street-view camera and a 3D point $(\theta, \phi, z)$ in the target coordinates increases as $z$ increases for $z>0$ and decreases as $z$ increases for $z<0$.
Figure~\ref{viewing_ray} provides an intuitive illustration.
The projected points from plane $z3$ will be nearer than those from planes $z1$ and $z2$, and the projected points from plane $z4$ will be nearer than those from planes $z5$ and $z6$.
All points with $z<0$ will be projected to the bottom half of the street-view panorama and those with $z>0$ will be imaged to the top half.
Therefore, simply adjusting the order of projected planes will help to sort points from far to near along viewing directions, as shown in Figure~\ref{plane_reorder}.
After that, Equation~\ref{alpha_composite} can be adopted to render the projected street-view panorama.
Note that the projected volumetric scene representation in Figure~\ref{plane_reorder} is not an MPI since the depth of image planes are not uniformly sampled.

We denote this projection as ``height-wise'' and our MPI-to-MPI projection as ``depth-wise''. 
Furthermore, the projected volumetric scene representation, the street-view MPI in Figure~\ref{fig:geometric_transfer} and the reordered planes in Figure~\ref{plane_reorder}, can be employed directly as an input to the generator instead of the rendered image.
Therefore, we investigate the performance of different satellite to street-view image projection methods with different network conditions.
The results
are presented in Table~\ref{ablation_a2g}. 
As indicated by the results, there is negligible difference in performance between the different approaches, while all of them outperforms existing methods (results presented in Table~\ref{tab: quantitative_stoa}).
This demonstrates that establishing the geometric correspondences between the two view images is indeed useful for the satellite to street-view image synthesis.
For the difference among different projection methods, using volumetric generator inputs requires slightly more model parameters and longer training time.
For the sake of better interpretability, we select the ``depth-wise'' projection and use the rendered image as the generator input.

{
\color{black}
\section{Discussion and Limitations}

\begin{figure}[t!]
\setlength{\abovecaptionskip}{0pt}
\setlength{\belowcaptionskip}{0pt}
    \centering
    \begin{minipage}{0.16\linewidth}
    \includegraphics[width=\linewidth]{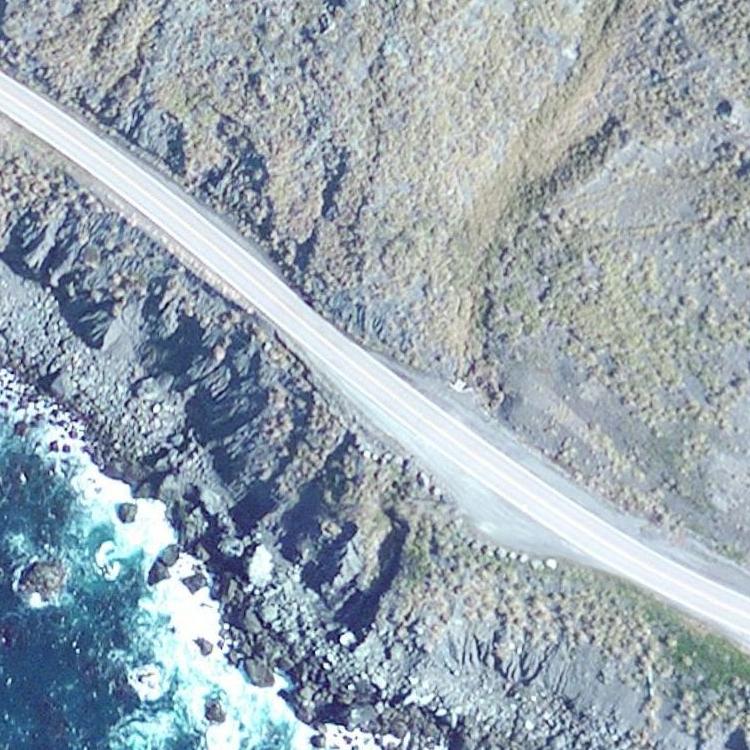}
    \includegraphics[width=\linewidth]{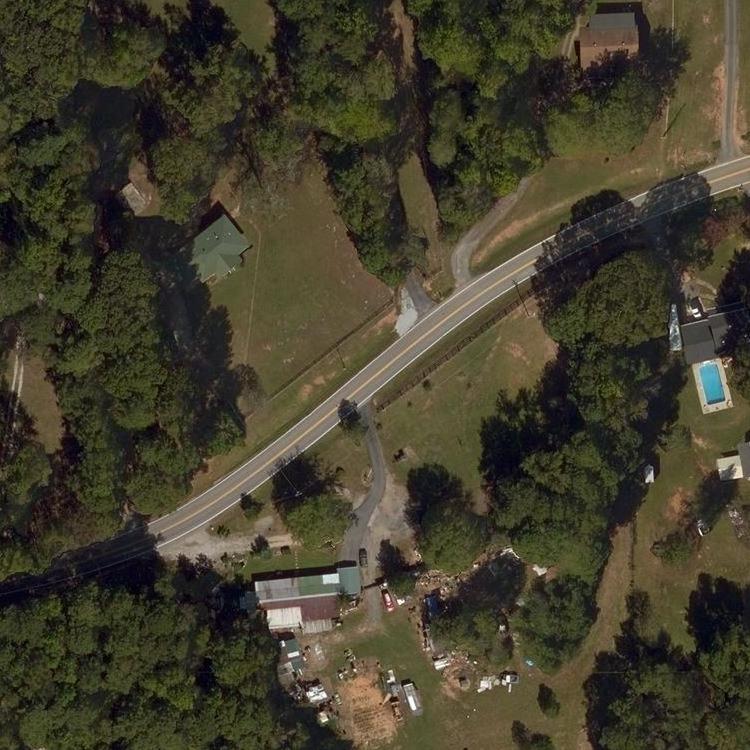}
    \centerline{\footnotesize Satellite Image}
    \end{minipage}
    \begin{minipage}{0.4\linewidth}
    \includegraphics[width=\linewidth, height=0.4\linewidth]{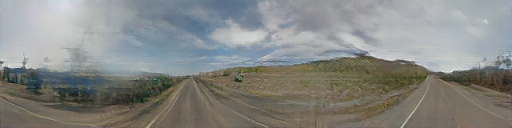}
    \includegraphics[width=\linewidth, height=0.4\linewidth]{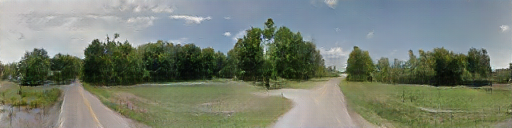}
    \centerline{\footnotesize Ours}
    \end{minipage}
     \begin{minipage}{0.4\linewidth}
    \includegraphics[width=\linewidth, height=0.4\linewidth]{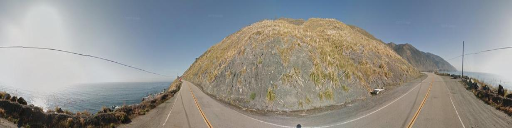}
    \includegraphics[width=\linewidth, height=0.4\linewidth]{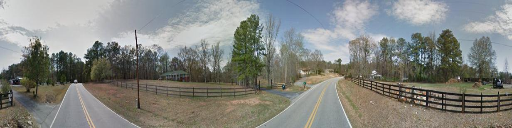}
    \centerline{\footnotesize Ground Truth}
    \end{minipage}
    \caption{\small Failure cases. 
    First row: the height of the hill is hard to estimate accurately from a single satellite image.
    Second row: the fence is difficult to be visible from above and is ignored.}
    \label{fig:failure_cases}
\end{figure}

For image generation tasks, skip connections between the encoder and decoder of a UNet-type generator can be useful for recovering fine detail.
However, their presence or absence made no difference to the performance of our method and the baseline methods (\ie, Pix2Pix and XFork).
For the baselines, this may be due to the significant differences in satellite and street-view image modalities and resolutions.
Our approach partially alleviates these differences with the S2SP module, but skip connections contribute to the final performance marginal.

The main limitation of our method is that it hallucinates the fa\c{c}ades of objects since a satellite image only views their top surfaces.
This can be seen in the trees of the second row of Figure~\ref{fig:visualization on CVUSA}.
In fact, this is a common limitation for all the existing satellite to street-view synthesis approaches.
Given a satellite image, there are many possibilities regarding the street-view appearance of objects (\eg, various building fa\c{c}ade structures, different colors, and different seasons).
It would be interesting to reformulate the task as a one-to-many problem, with a latent vector encoding the desired property in the target view image. 
Our method also fails when the heights are misestimated, such as the hill in the first row being taller than expected, and when objects are too small (indistinguishable) in the satellite images, such as the fence in the second row. 

Another limitation is the rendering speed: our approach takes 0.2s to synthesize one image, whereas Pix2Pix and XFork only take 0.02s. For time-critical applications, the baseline methods (Pix2Pix and XFork) could be used when the distribution of cross-view image pairs is uniform (\eg, the variation of scene structures at different locations is small), since the geometric correspondences can be learned statistically during training. However, our model (with slower rendering) is needed when the variation of scene structures is large, since the inductive bias encoded in our framework makes it easier for the network to learn the significant cross-view transformations.
}

\section{Conclusion}

In this paper, we have proposed a novel geometry-aware method for synthesizing street-view panoramas from satellite images. 
In contrast to existing methods that adopt black-box training procedures, our algorithm explicitly establishes the geometric correspondences between satellite and street-view images. 
The central innovation of this paper is the novel and differentiable satellite to street-view image projection module, which exploits the two-view geometry of the setup for accurate and occlusion-aware image synthesis.
Quantitative and qualitative experimental results demonstrate that our method is able to generate geometrically-consistent street-view panoramas from satellite images.
We expect the key idea of this paper (explicitly exposing the geometric correspondences and implicitly estimating the height map) to be useful for solving other related problems, such as estimating the height/depth maps of satellite/street-view images and novel view synthesis given two-view image pairs.


% % you can choose not to have a title for an appendix
% % if you want by leaving the argument blank
% \section{}

\ifCLASSOPTIONcompsoc
  \section*{Acknowledgments}
\else
  \section*{Acknowledgment}
\fi

% \section{ Acknowledgments}
This research is funded in part by the ARC-Discovery (DP 190102261).
Y.S. is a China Scholarship Council (CSC)-funded PhD student to ANU, and D.C. is grateful for support from Continental AG.
We thank AE and all reviewers for their time and patience in reviewing this paper and their constructive suggestions.

\ifCLASSOPTIONcaptionsoff
  \newpage
\fi



\bibliographystyle{IEEEtran}
\bibliography{egbib}

\begin{thebibliography}{10}
\providecommand{\url}[1]{#1}
\csname url@samestyle\endcsname
\providecommand{\newblock}{\relax}
\providecommand{\bibinfo}[2]{#2}
\providecommand{\BIBentrySTDinterwordspacing}{\spaceskip=0pt\relax}
\providecommand{\BIBentryALTinterwordstretchfactor}{4}
\providecommand{\BIBentryALTinterwordspacing}{\spaceskip=\fontdimen2\font plus
\BIBentryALTinterwordstretchfactor\fontdimen3\font minus
  \fontdimen4\font\relax}
\providecommand{\BIBforeignlanguage}[2]{{%
\expandafter\ifx\csname l@#1\endcsname\relax
\typeout{** WARNING: IEEEtran.bst: No hyphenation pattern has been}%
\typeout{** loaded for the language `#1'. Using the pattern for}%
\typeout{** the default language instead.}%
\else
\language=\csname l@#1\endcsname
\fi
#2}}
\providecommand{\BIBdecl}{\relax}
\BIBdecl

\bibitem{Regmi_2019_ICCV}
K.~Regmi and M.~Shah, ``Bridging the domain gap for ground-to-aerial image
  matching,'' in \emph{The IEEE International Conference on Computer Vision
  (ICCV)}, October 2019.

\bibitem{toker2021coming}
A.~Toker, Q.~Zhou, M.~Maximov, and L.~Leal-Taix{\'e}, ``Coming down to earth:
  Satellite-to-street view synthesis for geo-localization,'' \emph{arXiv
  preprint arXiv:2103.06818}, 2021.

\bibitem{porter1984compositing}
T.~Porter and T.~Duff, ``Compositing digital images in siggraph comput,''
  \emph{Graph}, vol.~18, no.~3, pp. 253--259, 1984.

\bibitem{lu2020geometry}
X.~Lu, Z.~Li, Z.~Cui, M.~R. Oswald, M.~Pollefeys, and R.~Qin, ``Geometry-aware
  satellite-to-ground image synthesis for urban areas,'' in \emph{Proceedings
  of the IEEE/CVF Conference on Computer Vision and Pattern Recognition}, 2020,
  pp. 859--867.

\bibitem{liu2018geometry}
M.~Liu, X.~He, and M.~Salzmann, ``Geometry-aware deep network for single-image
  novel view synthesis,'' in \emph{Proceedings of the IEEE Conference on
  Computer Vision and Pattern Recognition}, 2018, pp. 4616--4624.

\bibitem{zhou2018stereo}
T.~Zhou, R.~Tucker, J.~Flynn, G.~Fyffe, and N.~Snavely, ``Stereo magnification:
  learning view synthesis using multiplane images,'' \emph{ACM Transactions on
  Graphics (TOG)}, vol.~37, no.~4, pp. 1--12, 2018.

\bibitem{flynn2019deepview}
J.~Flynn, M.~Broxton, P.~Debevec, M.~DuVall, G.~Fyffe, R.~Overbeck, N.~Snavely,
  and R.~Tucker, ``Deepview: View synthesis with learned gradient descent,'' in
  \emph{Proceedings of the IEEE Conference on Computer Vision and Pattern
  Recognition}, 2019, pp. 2367--2376.

\bibitem{tucker2020single}
R.~Tucker and N.~Snavely, ``Single-view view synthesis with multiplane
  images,'' in \emph{Proceedings of the IEEE/CVF Conference on Computer Vision
  and Pattern Recognition}, 2020, pp. 551--560.

\bibitem{zhai2017predicting}
M.~Zhai, Z.~Bessinger, S.~Workman, and N.~Jacobs, ``Predicting ground-level
  scene layout from aerial imagery,'' in \emph{IEEE Conference on Computer
  Vision and Pattern Recognition}, vol.~3, 2017.

\bibitem{regmi2018cross}
K.~Regmi and A.~Borji, ``Cross-view image synthesis using conditional gans,''
  in \emph{Proceedings of the IEEE Conference on Computer Vision and Pattern
  Recognition}, 2018, pp. 3501--3510.

\bibitem{regmi2019crossview}
------, ``Cross-view image synthesis using geometry-guided conditional gans,''
  \emph{Computer Vision and Image Understanding}, vol. 187, p. 102788, 2019.

\bibitem{tang2019multi}
H.~Tang, D.~Xu, N.~Sebe, Y.~Wang, J.~J. Corso, and Y.~Yan, ``Multi-channel
  attention selection gan with cascaded semantic guidance for cross-view image
  translation,'' in \emph{Proceedings of the IEEE Conference on Computer Vision
  and Pattern Recognition}, 2019, pp. 2417--2426.

\bibitem{workman2015wide}
S.~Workman, R.~Souvenir, and N.~Jacobs, ``Wide-area image geolocalization with
  aerial reference imagery,'' in \emph{Proceedings of the IEEE International
  Conference on Computer Vision}, 2015, pp. 3961--3969.

\bibitem{workman2015location}
S.~Workman and N.~Jacobs, ``On the location dependence of convolutional neural
  network features,'' in \emph{Proceedings of the IEEE Conference on Computer
  Vision and Pattern Recognition Workshops}, 2015, pp. 70--78.

\bibitem{vo2016localizing}
N.~N. Vo and J.~Hays, ``Localizing and orienting street views using overhead
  imagery,'' in \emph{European Conference on Computer Vision}.\hskip 1em plus
  0.5em minus 0.4em\relax Springer, 2016, pp. 494--509.

\bibitem{Hu_2018_CVPR}
S.~Hu, M.~Feng, R.~M.~H. Nguyen, and G.~Hee~Lee, ``Cvm-net: Cross-view matching
  network for image-based ground-to-aerial geo-localization,'' in \emph{The
  IEEE Conference on Computer Vision and Pattern Recognition (CVPR)}, June
  2018.

\bibitem{sun2019geocapsnet}
B.~Sun, C.~Chen, Y.~Zhu, and J.~Jiang, ``Geocapsnet: Aerial to ground view
  image geo-localization using capsule network,'' \emph{arXiv preprint
  arXiv:1904.06281}, 2019.

\bibitem{Cai_2019_ICCV}
S.~Cai, Y.~Guo, S.~Khan, J.~Hu, and G.~Wen, ``Ground-to-aerial image
  geo-localization with a hard exemplar reweighting triplet loss,'' in
  \emph{The IEEE International Conference on Computer Vision (ICCV)}, October
  2019.

\bibitem{Liu_2019_CVPR}
L.~Liu and H.~Li, ``Lending orientation to neural networks for cross-view
  geo-localization,'' in \emph{The IEEE Conference on Computer Vision and
  Pattern Recognition (CVPR)}, June 2019.

\bibitem{shi2020optimal}
Y.~Shi, X.~Yu, L.~Liu, T.~Zhang, and H.~Li, ``Optimal feature transport for
  cross-view image geo-localization.'' in \emph{AAAI}, 2020, pp.
  11\,990--11\,997.

\bibitem{shi2019spatial}
Y.~Shi, L.~Liu, X.~Yu, and H.~Li, ``Spatial-aware feature aggregation for image
  based cross-view geo-localization,'' in \emph{Advances in Neural Information
  Processing Systems}, 2019, pp. 10\,090--10\,100.

\bibitem{shi2020looking}
Y.~Shi, X.~Yu, D.~Campbell, and H.~Li, ``Where am i looking at? joint location
  and orientation estimation by cross-view matching,'' \emph{arXiv preprint
  arXiv:2005.03860}, 2020.

\bibitem{isola2017image}
P.~Isola, J.-Y. Zhu, T.~Zhou, and A.~A. Efros, ``Image-to-image translation
  with conditional adversarial networks,'' in \emph{Proceedings of the IEEE
  conference on computer vision and pattern recognition}, 2017, pp. 1125--1134.

\bibitem{Simonyan2014VeryDC}
K.~Simonyan and A.~Zisserman, ``Very deep convolutional networks for
  large-scale image recognition,'' \emph{CoRR}, vol. abs/1409.1556, 2014.

\bibitem{kingma2015adam}
D.~P. Kingma and J.~Ba, ``Adam: A methodfor stochastic optimization,'' in
  \emph{International Conference onLearning Representations (ICLR)}, 2015.

\bibitem{zhang2018unreasonable}
R.~Zhang, P.~Isola, A.~A. Efros, E.~Shechtman, and O.~Wang, ``The unreasonable
  effectiveness of deep features as a perceptual metric,'' in \emph{Proceedings
  of the IEEE Conference on Computer Vision and Pattern Recognition}, 2018, pp.
  586--595.

\bibitem{krizhevsky2012imagenet}
A.~Krizhevsky, I.~Sutskever, and G.~E. Hinton, ``Imagenet classification with
  deep convolutional neural networks,'' in \emph{Advances in neural information
  processing systems}, 2012, pp. 1097--1105.

\bibitem{iandola2016squeezenet}
F.~N. Iandola, S.~Han, M.~W. Moskewicz, K.~Ashraf, W.~J. Dally, and K.~Keutzer,
  ``Squeezenet: Alexnet-level accuracy with 50x fewer parameters and< 0.5 mb
  model size,'' \emph{arXiv preprint arXiv:1602.07360}, 2016.

\bibitem{cordts2016cityscapes}
M.~Cordts, M.~Omran, S.~Ramos, T.~Rehfeld, M.~Enzweiler, R.~Benenson,
  U.~Franke, S.~Roth, and B.~Schiele, ``The cityscapes dataset for semantic
  urban scene understanding,'' in \emph{Proceedings of the IEEE conference on
  computer vision and pattern recognition}, 2016, pp. 3213--3223.

\bibitem{nekrasov2018light}
V.~Nekrasov, C.~Shen, and I.~Reid, ``Light-weight refinenet for real-time
  semantic segmentation,'' \emph{arXiv preprint arXiv:1810.03272}, 2018.

\bibitem{long2015fully}
J.~Long, E.~Shelhamer, and T.~Darrell, ``Fully convolutional networks for
  semantic segmentation,'' in \emph{Proceedings of the IEEE conference on
  computer vision and pattern recognition}, 2015, pp. 3431--3440.

\end{thebibliography}


% \end{thebibliography}


\begin{IEEEbiography}[{\includegraphics[width=1in,height=1.25in,clip,keepaspectratio]{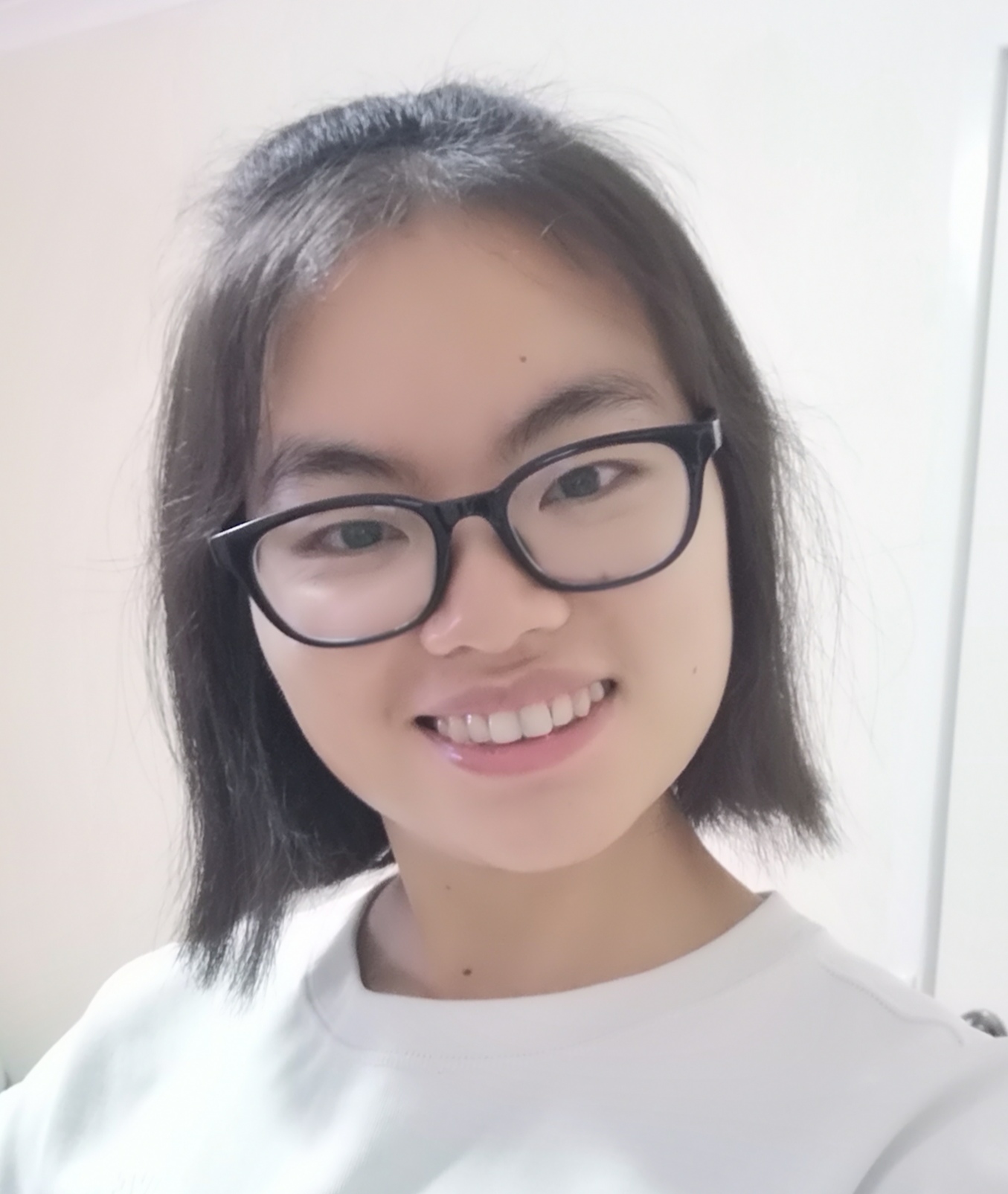}}]{Yujiao Shi}  is a PhD student at the College of Engineering and Computer Science (CECS), Australian National University. She received her B.E. degree and M.S. degree in automation from Nanjing University of Posts and Telecommunications, Nanjing, China, in 2014 and 2017, respectively. Her research interests include satellite image based geo-localization, novel view synthesis and scene understanding.
\end{IEEEbiography}

\begin{IEEEbiography}[{\includegraphics[width=1in,height=1.25in,clip,keepaspectratio]{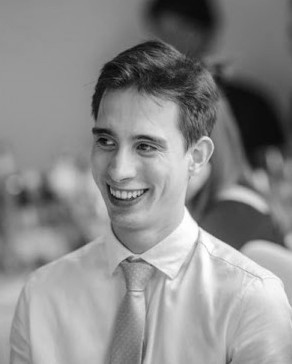}}]{Dylan Campbell}
is a Research Fellow with the Visual Geometry Group (VGG) at the University of Oxford. Prior to that he was a Research Fellow in the Research School of Computer Science at the Australian National University and the Australian Research Council Centre of Excellence in Robotic Vision. He received his PhD degree in Engineering from the Australian National University in 2018 while concurrently working as a research assistant in the Cyber-Physical Systems group at Data61–CSIRO. Prior to that, Dylan received a BE in Mechatronic Engineering from the University of New South Wales. His research interests cover a range of computer vision and machine learning topics, including visual geometry and differentiable optimization.
\end{IEEEbiography}

\begin{IEEEbiography}[{\includegraphics[width=1in,height=1.25in,clip,keepaspectratio]{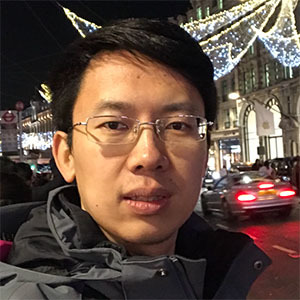}}]{Xin Yu}
received his B.S. degree in Electronic Engineering from University of Electronic Science and Technology of China, Chengdu, China, in 2009, and received his Ph.D. degree in the Department of Electronic Engineering, Tsinghua University, Beijing, China, in 2015. He also received a Ph.D. degree in the College of Engineering and Computer Science, Australian National University, Canberra, Australia, in 2019. He is currently a senior lecturer in University of Technology Sydney. His interests include computer vision and image processing.
\end{IEEEbiography}

\begin{IEEEbiography}[{\includegraphics[width=1in,height=1.25in,clip,keepaspectratio]{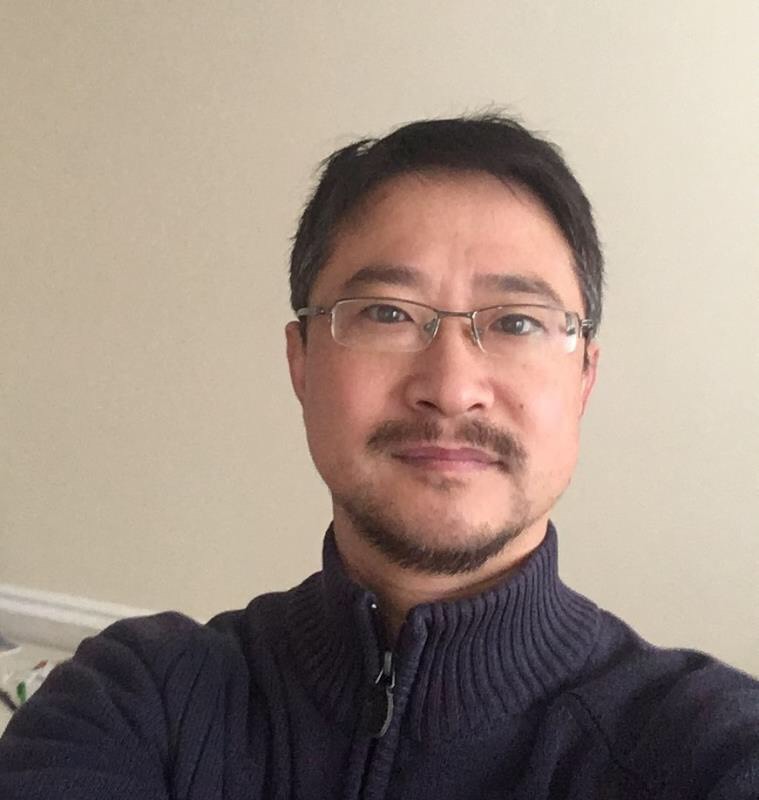}}]{Hongdong Li} is a Professor of ANU.  He is also a founding Chief Investigator for the Australia Centre of Excellence for Robotic Vision (ACRV). His research interests include 3D vision reconstruction, structure from motion, multi-view geometry, as well as applications of optimization methods in computer vision.  Prior to 2010, he was with NICTA working on the “Australia Bionic Eyes” project.  He is an Associate Editor for IEEE T-PAMI, Guest editor for IJCV, and Area Chair in recent year ICCV, ECCV and CVPR conferences.  He was a Program Chair for ACRA 2015 – Australia Conference on Robotics and Automation, and a Program Co-Chair for ACCV 2018 – Asian Conference on Computer Vision.  He won a number of paper awards in computer vision and pattern recognition, and was the receipt for the CVPR 2012 Best Paper Award, the ICCV Marr Prize Honorable Mention in 2017, and a shortlist of the CVPR 2020 best paper award.
\end{IEEEbiography}

% \end{IEEEbiography}


% % insert where needed to balance the two columns on the last page with
% %\newpage



%\vfill


\end{document}